\documentclass[10pt,journal,twoside,compsoc]{IEEEtran}

\usepackage{amsmath}
\usepackage{graphicx}
\usepackage{txfonts}
\usepackage{multirow}
\usepackage{verbatim}
\usepackage{array}
\usepackage{subfig}
\usepackage[usenames, dvipsnames]{xcolor}
\usepackage[nocompress]{cite}
\usepackage[colorlinks, linkcolor=black, urlcolor = black, citecolor=black]{hyperref}
\usepackage{paperSyml}
\usepackage{pifont}
\usepackage{paperSyml}

\newcommand{\YES}{\color{ForestGreen}{\ding{52}}}
\newcommand{\NO}{\color{Red}{\ding{56}}}

\begin{document}

\title{Location-Centered House Price Prediction: A Multi-Task Learning Approach}

\author{
Guangliang~Gao,
\IEEEcompsocitemizethanks{
	\IEEEcompsocthanksitem G. Gao is with the School of Computer Science and Engineering, Nanjing University of Science and Technology, Nanjing, China.\protect\\
	E-mail: guangliang.gao@njust.edu.cn}
Zhifeng~Bao,
\IEEEcompsocitemizethanks{
	\IEEEcompsocthanksitem Z. Bao is with the School of Science, Computer Science, and Information Technology, RMIT University, Melbourne, VIC 3000, Australia.\protect\\
	E-mail: zhifeng.bao@rmit.edu.au}
Jie~Cao,
\IEEEcompsocitemizethanks{
	\IEEEcompsocthanksitem J. Cao is with the School of Information Engineering, Nanjing University of Finance and Economics, Nanjing, China, and with the College of Computer Science and Engineering, Nanjing University of Science and Technology, Nanjing, China.\protect\\
	E-mail: caojie690929@163.com}
A. K.~Qin,
Timos~Sellis,~\IEEEmembership{Fellow,~IEEE,}
\IEEEcompsocitemizethanks{
	\IEEEcompsocthanksitem A. K. Qin and T. Sellis are with the Department of Computer Science and Software Engineering, Swinburne University of Technology, Melbourne, VIC 3122, Australia.\protect\\
	E-mail: \{kqin, tsellis\}@swin.edu.au.}
Zhiang~Wu
\IEEEcompsocitemizethanks{
	\IEEEcompsocthanksitem Z. Wu (corresponding author) is with the School of Information Engineering, Nanjing University of Finance and Economics, Nanjing, China.\protect\\
	E-mail: zawuster@gmail.com}
}

\IEEEtitleabstractindextext{%
\begin{abstract}
Accurate house prediction is of great significance to various real estate stakeholders such as house owners, buyers, investors, and agents. We propose a location-centered prediction framework that differs from existing work in terms of data profiling and prediction model. Regarding data profiling, we make an important observation as follows -- besides the in-house features such as floor area, the location plays a critical role in house price prediction. Unfortunately, existing work either overlooked it or had a coarse grained measurement of locations. Thereby, we define and capture a fine-grained location profile powered by a diverse range of location data sources, such as transportation profile (e.g., distance to nearest train station, door-to-door travel time by public transportation), education profile (e.g., school zones and ranking), suburb profile based on census data, facility profile (e.g., nearby GPs, hospitals, supermarkets). 
Regarding the choice of prediction model, we observe that a variety of approaches either consider the entire data for modeling, or split the entire house data and model each partition independently. However, such modeling ignores the relatedness between partitions, and for all prediction scenarios, there may not be sufficient training samples per partition for the latter approach. We address this problem by conducting a careful study of exploiting the Multi-Task Learning (MTL) model. Specifically, we map the strategies for splitting the entire house data to the ways the tasks are defined in MTL, and each partition obtained is aligned with a task. Furthermore, we select specific MTL-based methods with different regularization terms to capture and exploit the relatedness between tasks. Based on real-world house transaction data collected in Melbourne, Australia from Jan. 2015 to Jan. 2018, these transactions are recorded daily for three years, including house sales price and various house features. We design extensive experimental evaluations, and the results indicate a significant superiority of MTL-based methods over state-of-the-art approaches. Meanwhile, we conduct an in-depth analysis on the impact of task definitions and method selections in MTL on the prediction performance, and demonstrate that the impact of task definitions on prediction performance far exceeds that of method selections.
\end{abstract}

\begin{IEEEkeywords}
Price prediction, real estate, multi-task learning
\end{IEEEkeywords}
}
\maketitle

\IEEEraisesectionheading{\section{Introduction}\label{sect1}}
\IEEEPARstart{W}{ith} the improvement of people's living standards, the demand for houses increases. In the United States{\footnote{\url{https://www.statista.com/statistics/226144/us-existing-home-sales/}}}, house sales have grown by 34\% in the last decade and reached a record high of 5.51 million last year. In Australia{\footnote{\url{https://tradingeconomics.com/australia/new-home-sales}}}, house sales have increased by 36\% since 2013. House price prediction~\cite{Kain1970, schulz2004state} has therefore attracted widespread attentions because the prediction outcomes can help various real estate stakeholders to make more informed decisions. For example, buyers would use house price prediction to search for candidate houses that match their financial capabilities. Similarly, house owners would need it to keep monitoring the market and seek the best opportunity for house selling. Moreover, real estate sales agents also rely on house price prediction to help customers better find out market trends, and the accuracy of prediction has become an important criterion for measuring the credibility of house sales agents.

House price is considered to be related to various house features. In general, these features can be grouped into two categories: non-geographical features, such as the number of bedrooms and floor space area; geographical features, such as the distance to the city center and the quality of nearby schools. Therefore, research on house price prediction defines appropriate models to fit various features to predict house price. The hedonic price model~\cite{10.2307/1828835, rosen1974hedonic} proposed from the perspective of economics is the most typical representative, and has been studied extensively in the literature of house price prediction~\cite{can1992specification, krol2015application, yayar2014hedonic}. However, it is primarily used for analyzing the relationship between house price and house features, where it typically adopts regression methods. In recent years, with the extensive application of machine learning in various fields, house price prediction through more machine learning methods, such as ANN (Artificial Neural Network)~\cite{selim2009determinants}, SVM (Support Vector Machine)~\cite{gu2011housing, wang2014real}, AdaBoost (Adaptive Boosting)~\cite{DBLP:journals/eswa/ParkB15}, has also received more and more attention. 

By carefully examining these studies, it reveals that modeling is usually carried out from a global view, i.e., implemented on the entire house data directly. However, the prediction performance of such approaches is not satisfactory, especially as the scale of the house data increases. The primary reason is that the weight of various house features is always assumed to be constant in the formulation of house price from the global view. In fact, however, the impact of house features on house price varies from house to house. For example, transportation features may have a greater impact on house price in the suburbs than that in urban centers, the difference in house price between school districts and non-school districts will be mainly concentrated on education features. 

Numerous studies~\cite{bourassa2010predicting, case2004modeling, GEREK201433, Montero2018} have also demonstrated that the different locations of houses and the surrounding communities of houses have a significant impact on the price of houses, though at a very coarse granularity. As a consequence, the researchers began splitting the entire house data into several partitions and designing prediction models for each partition individually. However, no matter which strategy is used for splitting, there are still two challenges that limit further improvements in the prediction performance along this line: (1) independent modeling ignores the relatedness among the partitions; (2) each model uses only the data of its corresponding partition. Data sparsity, especially in partitions where the number of original data is insufficiently small, is a serious problem.

To address the aforementioned issues, we exploit the framework of Multi-Task Learning (MTL) to model the house price prediction problem. MTL~\cite{argyriou2007multi, Caruana1997, evgeniou2004regularized, zhang2017survey} is a type of transfer learning, when there are multiple related tasks and each task has limited training samples, the model can enhance the performance by extracting and utilizing shared information among different tasks. MTL has been applied in many applications, including time series analysis~\cite{chidlovskii2017multi}, stock selection~\cite{ghosn1997multi}, event forecasting~\cite{jaques2016multi, zhao2015multi}, disease progression~\cite{emrani2017prognosis, zhou2011multi}, and water quality prediction~\cite{liu2016urban}. By using MTL, we  inherit the principle of modeling house price prediction problem from local views because the entire house data is grouped into multiple tasks. Moreover, MTL considers the relatedness among tasks, and the shortcomings of insufficient samples and independent modeling can also be addressed.

There are two key points in MTL: (i) how to define tasks and (ii) how to characterize the relatedness among tasks. In general, the first point is data-driven, different strategies for defining tasks will result in different task sets, and the relatedness among tasks will also be determined. We split the entire house data and define each partition as a task after splitting, where the splitting strategy is equivalent to the strategy for defining tasks in MTL. The second point is method-driven -- once a multi-task problem is formulated, the design of a specific method will indicate the level of learning. We select MTL-based methods with regularization terms to capture and utilize the relatedness among tasks, and different regularization terms represent different levels of learning. To this end, in this paper, we present an organized study on exploiting MTL for the problem of house price prediction. The specific contributions of our work are listed as follows:

\begin{itemize}
\item{We formulate house price prediction as an MTL problem. To the best of our knowledge, this work is the first to consider using MTL to solve house price prediction problem, which provides a new perspective on the study of house price prediction. Additionally, our work also enriches the application fields of MTL.}

\item{We define and capture a fine-grained location profile powered by a diverse range of location data sources. We observe that the location of house plays a critical role in house price prediction. Therefore, we focus on enriching the location-driven house features and grouping them into four profiles for further fine-grained, namely house, education, transportation and facility, respectively.}

\item{We demonstrate the superiority of MTL-based methods over state-of-the-art approaches on real-world house data. Based on our house data, we evaluate the prediction performance of MTL-based methods and five state-of-the-art approaches. Experimental results show that MTL-based methods consistently outperform these competing approaches.}

\item{We conduct an in-depth analysis on the impact of task definitions and method selections on prediction performance. We design two categories of strategies to define tasks and select three MTL-based methods with different regularization terms. By comparing their corresponding prediction performance, we reveal that the impact of task definitions on prediction performance far exceeds that of method selections.}
\end{itemize}

The remainder of this paper is organized as follows. We review the related work in Section~\ref{sect2}, and Section~\ref{sect3} provides a comprehensive profiling of our location-centered house data. The MTL-based house price prediction is described in Section~\ref{sect4}, and Section~\ref{sect5} shows our experimental results. Finally, we conclude the paper and give guidelines for our future work in Section~\ref{sect6}.

\section{Related Work}\label{sect2}
\begin{table*}[t]
\centering
\begin{tabular}{| p{1.5cm} | p{2.3cm} | c | c | c | c | c | c | c | c | c | c | c | c | c | c | c | c |}
\hline
	&\multirow{2}{2.3cm}{Data Examples}	&\multicolumn{14}{c |}{References}	&\multirow{2}{*}{Ours}\\
\cline{3-16}
	&	&\cite{can1992specification}	&\cite{krol2015application}	&\cite{yayar2014hedonic}		&\cite{selim2009determinants}	&\cite{DBLP:journals/eswa/ParkB15}	&\cite{bourassa2010predicting}	&\cite{case2004modeling}	&\cite{Montero2018}	&\cite{fan2006determinants}	&\cite{kryvobokov2007analysing}	&\cite{Ottensmann2008Urban}	&\cite{ozalp2017use}	&\cite{kuntz2014geostatistical}	&\cite{adair1996hedonic}	&\\
\hline
\multirow{4}{1.4cm}{Scale \\of data}
&100		&{\YES}	&{\NO}	&{\NO}	&{\NO}	&{\NO}	&{\NO}	&{\NO}	&{\NO}	&{\NO}	&{\NO}	&{\YES}	&{\NO}	&{\YES}	&{\NO}	&{\NO}\\
\cline{2-2}
&1, 000	&{\NO}	&{\NO}	&{\YES}	&{\YES}	&{\NO}	&{\NO}	&{\NO}	&{\YES}	&{\NO}	&{\YES}	&{\NO}	&{\YES}	&{\NO}	&{\YES}	&{\NO}\\
\cline{2-2}
&10, 000	&{\NO}	&{\YES}	&{\NO}	&{\NO}	&{\YES}	&{\YES}	&{\YES}	&{\NO}	&{\YES}	&{\NO}	&{\NO}	&{\NO}	&{\NO}	&{\NO}	&{\NO}\\
\cline{2-2}
&100, 000	&{\NO}	&{\NO}	&{\NO}	&{\NO}	&{\NO}	&{\NO}	&{\NO}	&{\NO}	&{\NO}	&{\NO}	&{\NO}	&{\NO}	&{\NO}	&{\NO}	&{\YES}\\
\hline
\multirow{6}{1.4cm}{House\\profile}
&floor area, number of bedrooms	&\multirow{2}{*}{\YES}	&\multirow{2}{*}{\YES}	&\multirow{2}{*}{\YES}	&\multirow{2}{*}{\YES}	&\multirow{2}{*}{\YES}	&\multirow{2}{*}{\YES}	&\multirow{2}{*}{\YES}	&\multirow{2}{*}{\YES}	&\multirow{2}{*}{\YES}	&\multirow{2}{*}{\YES}	&\multirow{2}{*}{\YES}	&\multirow{2}{*}{\YES}	&\multirow{2}{*}{\YES}	&\multirow{2}{*}{\YES}	&\multirow{2}{*}{\YES}\\
\cline{2-2}
&geo-information, address, suburb	&\multirow{2}{*}{\NO}	&\multirow{2}{*}{\NO}	&\multirow{2}{*}{\YES}	&\multirow{2}{*}{\YES}	&\multirow{2}{*}{\YES}	&\multirow{2}{*}{\YES}	&\multirow{2}{*}{\YES}	&\multirow{2}{*}{\NO}	&\multirow{2}{*}{\NO}	&\multirow{2}{*}{\YES}	&\multirow{2}{*}{\YES}	&\multirow{2}{*}{\NO}	&\multirow{2}{*}{\YES}	&\multirow{2}{*}{\YES}	&\multirow{2}{*}{\YES}\\
\cline{2-2}
&air condition, water, heating views	&\multirow{2}{*}{\YES}	&\multirow{2}{*}{\YES}	&\multirow{2}{*}{\YES}	&\multirow{2}{*}{\YES}	&\multirow{2}{*}{\YES}	&\multirow{2}{*}{\YES}	&\multirow{2}{*}{\YES}	&\multirow{2}{*}{\YES}	&\multirow{2}{*}{\YES}	&\multirow{2}{*}{\NO}	&\multirow{2}{*}{\NO}	&\multirow{2}{*}{\YES}	&\multirow{2}{*}{\NO}	&\multirow{2}{*}{\YES}	&\multirow{2}{*}{\YES}\\
\hline
\multirow{3}{1.4cm}{Education\\profile}
&nearby schools	&{\NO}	&{\NO}	&{\YES}	&{\NO}	&{\YES}	&{\NO}	&{\NO}	&{\NO}	&{\NO}	&{\NO}	&{\NO}	&{\NO}	&{\YES}	&{\NO}	&{\YES}\\
\cline{2-2}
&school districts	&{\NO}	&{\NO}	&{\NO}	&{\NO}	&{\NO}	&{\NO}	&{\NO}	&{\NO}	&{\NO}	&{\NO}	&{\NO}	&{\YES}	&{\NO}	&{\NO}	&{\YES}\\
\cline{2-2}
&school rankings	&{\NO}	&{\NO}	&{\NO}	&{\NO}	&{\YES}	&{\NO}	&{\NO}	&{\NO}	&{\NO}	&{\NO}	&{\NO}	&{\YES}	&{\NO}	&{\NO}	&{\YES}\\
\hline
\multirow{3}{1.4cm}{Transportation\\profile}
&nearby public transport	&\multirow{2}{*}{\NO}	&\multirow{2}{*}{\NO}	&\multirow{2}{*}{\YES}	&\multirow{2}{*}{\NO}	&\multirow{2}{*}{\NO}	&\multirow{2}{*}{\NO}	&\multirow{2}{*}{\NO}	&\multirow{2}{*}{\NO}	&\multirow{2}{*}{\NO}	&\multirow{2}{*}{\NO}	&\multirow{2}{*}{\YES}	&\multirow{2}{*}{\NO}	&\multirow{2}{*}{\YES}	&\multirow{2}{*}{\YES}	&\multirow{2}{*}{\YES}\\
\cline{2-2}
&travel time to work	&{\NO}	&{\YES}	&{\NO}	&{\NO}	&{\NO}	&{\NO}	&{\NO}	&{\NO}	&{\NO}	&{\NO}	&{\YES}	&{\YES}	&{\YES}	&{\NO}	&{\YES}\\
\hline
\multirow{3}{1.4cm}{Facility\\profile}
&hospitals, shops	&{\NO}	&{\NO}	&{\YES}	&{\NO}	&{\NO}	&{\NO}	&{\NO}	&{\NO}	&{\YES}	&{\YES}	&{\YES}	&{\YES}	&{\YES}	&{\NO}	&{\YES}\\
\cline{2-2}
&distance to nearest hospitals	&\multirow{2}{*}{\NO}	&\multirow{2}{*}{\NO}	&\multirow{2}{*}{\NO}	&\multirow{2}{*}{\NO}	&\multirow{2}{*}{\NO}	&\multirow{2}{*}{\NO}	&\multirow{2}{*}{\NO}	&\multirow{2}{*}{\NO}	&\multirow{2}{*}{\YES}	&\multirow{2}{*}{\NO}	&\multirow{2}{*}{\YES}	&\multirow{2}{*}{\YES}	&\multirow{2}{*}{\YES}	&\multirow{2}{*}{\NO}	&\multirow{2}{*}{\YES}\\
\hline
\end{tabular}
\caption{Summary of our data profiles and comparisons with most of the existing house price prediction work. Our location-centered house data is more comprehensive in terms of house transaction records and house features than those used in the literature.}
\label{tab:sect21}
\end{table*}

House is usually treated as a heterogeneous goods, defined by a bundle of utility bearing features~\cite{fan2006determinants, KUSAN20101808}. Therefore, the house price can be considered as a quantitative representation of a set of these features. Over the past decades, a large amount of studies have examined the relationship between house price and house features. For example, Kr{\'o}l~\cite{krol2015application} investigated the relationship between the price of an apartment and its significant features based on the results of hedonic analysis in Poland. The work of~\cite{yayar2014hedonic} discussed which house features have negative or positive effects upon the value of the house in Turkey. Kryvobokov and Wilhelmsson~\cite{kryvobokov2007analysing} derived the weights of the relative importance of location features that influence the market values of apartments in Donetsk, Ukraine. Ottensmann et al.~\cite{Ottensmann2008Urban} compared measures of location using both distances and travel time, to the CBD, and to multiple employment centers to understand how residence location relative to employment location affects house price in Indianapolis, Indiana, USA. Ozalp and Akinci~\cite{ozalp2017use} determined the housing and environmental features that were effective on residential real estate sale prices in Artvin, Turkey.

Based on a broad study of the relationship between house price and various house features, house price prediction approaches output the estimated house price by inputting house features. According to the basic idea whether it relies on the global model or not, one can divide the existing house price prediction methods into two categories.

The global model predicts the house price on a range of its constituent features, and is usually modeled directly across the entire house data. Much work has been done along this line. Selim~\cite{selim2009determinants} examined the determinants of house price in Turkey by using the hedonic model and demonstrated that artificial neural network can be a better alternative for prediction of the house price in Turkey. Gu et al.~\cite{gu2011housing} proposed a hybrid of genetic algorithm and support vector machine approach (G-SVM) to predict house price, the cases from China showed the prediction ability of the method. Wang et al.~\cite{wang2014real} proposed a novel model based on SVM to predict the average house price in different years, meanwhile, the authors demonstrated that PSO algorithm can effectively determine the parameters of SVM. Park and Bae~\cite{DBLP:journals/eswa/ParkB15} developed a general prediction model based on machine learning methods such as C4.5, RIPPER, Naive Bayesian, and AdaBoost and compared their classification accuracy performance. The reason why global modeling can be widely recognized in house price prediction is obvious, because it is easy to apply and can reveal the comparative size of effects of various features on house price. However, global modeling ignores the impact of house location and surroundings on house price, so the prediction performance is often unsatisfactory as the scale of the house data increases.

Recent studies have focused on house price prediction from local views and have gradually become a serious alternative and extension of conventional house price modeling approaches. Among these studies, Bourassa et al.~\cite{bourassa2010predicting} compared alternative methods for taking spatial dependence into account in house price prediction, and concluded that a geostatistical model with disaggregated submarket variables performed the best. Case et al.~\cite{case2004modeling} investigated the hedonic model and three spatial models, and out-of-sample prediction accuracy was used for comparison purposes. Their prediction results indicated the importance to incorporate the nearest neighbor transactions for predicting housing values. Gerek~\cite{GEREK201433} designed two different adaptive neuro-fuzzy approaches for prediction, namely ANFIS with grid partition (ANFIS-GP) and ANFIS with sub clustering (ANFIS-SC), and the results indicated that the performance of ANFIS-GP was slightly better than that of ANFIS-SC. Montero et al.~\cite{Montero2018} considered parametric and semi-parametric spatial hedonic model variants to reflect the spatial effects in house price. The proposed model is represented as a mixed model that account for spatial autocorrelation, spatial heterogeneity and (smooth and nonparametrically specified) nonlinearities using penalized splines methodology. The results obtained suggest that the nonlinear models are the best strategies for house price prediction.

Although the house price prediction problem has been widely studied, our work is significantly different from most of the existing work in the following aspects. First, as shown in Table~\ref{tab:sect21}, our house data (more details will be provided in the next section) is more adequate in terms of house transaction records and house features than the house data used in the literature, which enables us to better explore the impact of various house features on house price and study house price prediction problem. Second, the existing studies, whether global or local, can be summarized into traditional Single Task Learning (STL), while we use MTL to study house price prediction. To the best of our knowledge, our work is the first one adopting MTL for price prediction. In STL, each task is considered to be independent and learn individually. In MTL, tasks learn simultaneously by using relatedness between each other. Combined with our previous statements, the principle of MTL can help us better model house price prediction problem from a finer-grained local view.

\section{House Data Profiling}\label{sect3}
In this section, we elaborate a comprehensive real estate dataset used in this study. We first describe four location-centered profiles, each of which contain a variety of features. Then we analyze the house data to further demonstrate the motivations for introducing MTL to predict house prices. 

\subsection{Data description}
We utilize house data from Melbourne{\footnote{\url{http://www.realestate.com.au/}}}~\cite{10.1007/978-3-319-46922-5_34, LI20181}, one of the largest cities in Australia, as an example to understand the domain situation. The data includes houses sold in Melbourne's metropolitan area since 2011. We extract 136, 394 house transaction records from Jan. 2015 to Jan. 2018 to generate the dataset for this study. The house features are divided into four profiles to better observe the impact of different types of features on house price. In particular, this comprehensive dataset contains four location-centered profiles: house profile, education profile, transportation profile and facility profile. 

\noindent\textbf{House profile.} In this group, we choose seven relevant features about the house itself. The number of bedrooms, the number of bathrooms, and land size are the most common basic features. The number of parking spaces is gradually related to price with the booming house market, thus we introduce this feature. Considering the possible correlation between price and income, we also include family weekly income as an independent feature. Moreover, geographical information has a great impact on house price. In the vast majority of cases, consumers are more concerned with general locations than with detailed addresses, so we choose two regional features, {\SAID}{\footnote{\url{http://www.abs.gov.au/}}} (Statistical Area Level, the geographical areas for the processing and release of Australian census data) and postal code, to reflect the impact of geographical information. Table~\ref{tab:sect31} reports the number of partitions at different statistical area levels and postal code.
 
\begin{table}[!htbp]
\centering
\begin{tabular}{| c | c |}
\hline
Statistical area level	&\#Partitions\\
\hline
{\SAfourID}	&17\\
\hline
{\SAthreeID}	&65\\
\hline
{\SAtwoID}	&100\\
\hline
Postal code	&547\\
\hline
{\SAoneID}	&10703\\
\hline
\end{tabular}
\caption{Number of partitions at different statistical area levels. {\SAfourID} has the coarsest-grained split of an area and the smallest number of partitions, while {\SAoneID} has the finest-grained split of an area and the largest number of partitions.}
\label{tab:sect31}
\end{table} 
 
\noindent\textbf{Education profile.} In recent years, educational resources have received more and more attention, thus we either find the exact primary and secondary school districts{\footnote{\url{http://melbourneschoolzones.com/}}} (top 20\% schools) that each house belongs to, or map the house with its nearest primary and secondary schools. The corresponding school rankings{\footnote{\url{http://bettereducation.com.au/}}} as four features to examine the impact on house price.

\noindent\textbf{Transportation profile.} Since transportation networks have always been of great concern, we set up six features: (1) the distance and walking time from each house to its nearest train station using Google Maps API{\footnote{\url{http://developers.google.com/maps/}}}, (2) the distance and travel time between each pair of train stations based on the GTFS{\footnote{\url{http://www.data.vic.gov.au/data/dataset/}}} (General Transit Feed Specification) data, and (3) the distance and self-driving time from the location of the house to the city center, i.e., to the main Central Business District (CBD), using Google Maps API. 

\noindent\textbf{Facility profile.} Proximity to facilities such as shops, hospitals, clinics, and supermarkets may affect the house price as well. Therefore, we introduce four different features based upon these four facilities to describe the distance between a given house and the nearest four facilities, where the distance calculation uses Google Maps API. 

\begin{figure}[t]
\centering
\includegraphics[width=0.48\textwidth]{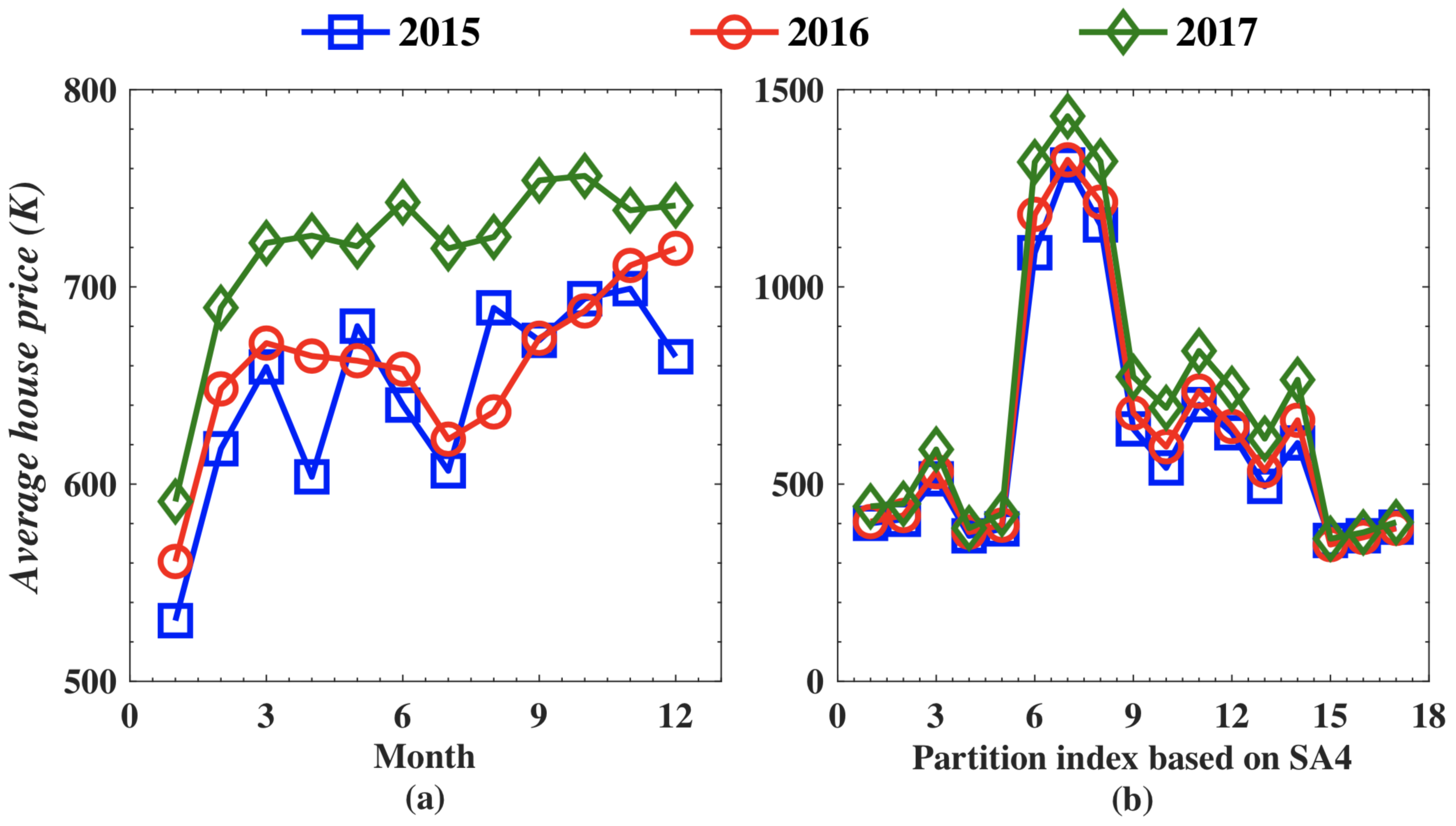}
\caption{Three-year average house price trend by Month and {\SAfourID}, respectively. Average house price fluctuates over time and varies by partition.}
\label{fig:sect31}
\end{figure}

Table~\ref{tab:sect32} summarizes all selected house features and their definitions. Meanwhile, some important statistics for our data set can be also found in Table~\ref{tab:sect32}. The value we want to predict, the house price at a given sales time, as the target. Figure~\ref{fig:sect31}a describes the trend of average house price in each month during the three years. It is clear that average house price is fluctuating over time, which indicates that house price is time-sensitive. Considering that location is one of the important factors in shaping the price of a house. Without loss of generality, we choose the partitions based on {\SAfourID} here. As shown in Figure~\ref{fig:sect31}b, the average house price in each partition is different. Therefore, spatial dependence also clearly exists in house price.

\begin{table*}[!htbp]
\small\centering
\begin{tabular}{| c | c | c | c | c | c | c |}
\hline
Category	&Name of features	&Descriptions	&Min.	&Max.	&Median	&Std. Dev.\\
\hline
\multirow{7}{*}{House}
&{\BEDROOM}	&The number of bedrooms	&1	&5	&--	&--\\
&{\BATHROOM}	&The number of bathrooms	&1	&3	&--	&--\\
&{\PARKING}	&The number of parking spaces	&1	&5	&--	&--\\
&{\LANDSIZE}	&The land size of the house ($m^2$)	&340		&2500	&708.79	&291.73\\ 
&{\INCOME}	&Family weekly income ($K$)	&935		&2836	&1553.91	&387.96\\
&{\SAID}	&Statistical area level	&SA1	&SA4	&--	&--\\
&{\POSTCODE}	&Postal code	&3000	&3996	&--	&--\\
\hline
\multirow{4}{*}{Education}
&{\SCHDIST}	&School district where the house is located	&1	&100		&--	&--\\
&{\NEARSCH}	&School closest to the house	&1	&500		&--	&--\\
&{\PRIRANK}	&The ranking of primary school	&3	&500		&--	&--\\
&{\SECRANK}	&The ranking of secondary school	&1	&500		&--	&--\\
\hline
\multirow{6}{*}{Transportation}
&{\DISTSTAT}	&Distance to the nearest train station ($m$)	&23	&5040	&2026.22	&1186.78\\
&{\TIMESTAT}	&Walking time to the nearest train station ($min$)	&1	&126		&34.75	&20.36\\
&{\DISTCBD}	&The train distance to city center ($m$)	&1300	&82600	&35239.52	&14594.42\\
&{\TIMECBD}	&The train time to city center ($min$)	&6	&101		&48.33	&16.64\\
&{\PDISTCBD}	&The self-driving distance to city center ($m$)	&1245	&83497	&35109.47	&14678.34\\
&{\PTIMECBD}	&The self-driving time to city center ($min$)	&10	&120		&47.98	&18.58\\
\hline
\multirow{4}{*}{Facility}
&{\DISTSHOP}	&The distance to nearest shopping center	($m$)	&5	&4999	&1643.19	&961.48\\
&{\DISTHOSP}	&The distance to nearest hospital ($m$)	&15	&5000	&1832.83	&1102.66\\
&{\DISTGP}	&The distance to nearest clinic ($m$)	&8	&4999	&957.21	&745.01\\
&{\DISTMARK}	&The distance to nearest supermarket ($m$)	&25	&5000	&1452.63	&847.71\\
\hline
\multicolumn{2}{ |c|}{\DATE}	&The sales time of the houses	&Jan. 2015	&Jan. 2018	&--	&--\\
\multicolumn{2}{ |c|}{\PRICE}	&The sales price of the houses ($K$)	&262		&2090	&680.54	&353.97\\
\hline
\end{tabular}
\caption{List of selected house features and statistics of our house data.}
\label{tab:sect32}
\end{table*}

\subsection{Data insight}
Because of the time sensitivity and spatial dependence of house price as described above, the researchers~\cite{bourassa2010predicting, case2004modeling, kuntz2014geostatistical, Montero2018} intuitively split house data and model each partition individually. However, such modeling usually does not deal well with house price prediction problem. The two challenges that affect the prediction performance have been mentioned in the previous sections, and we now elaborate on them by analyzing the house data.

\begin{figure}[t]
\centering
\includegraphics[width=0.48\textwidth]{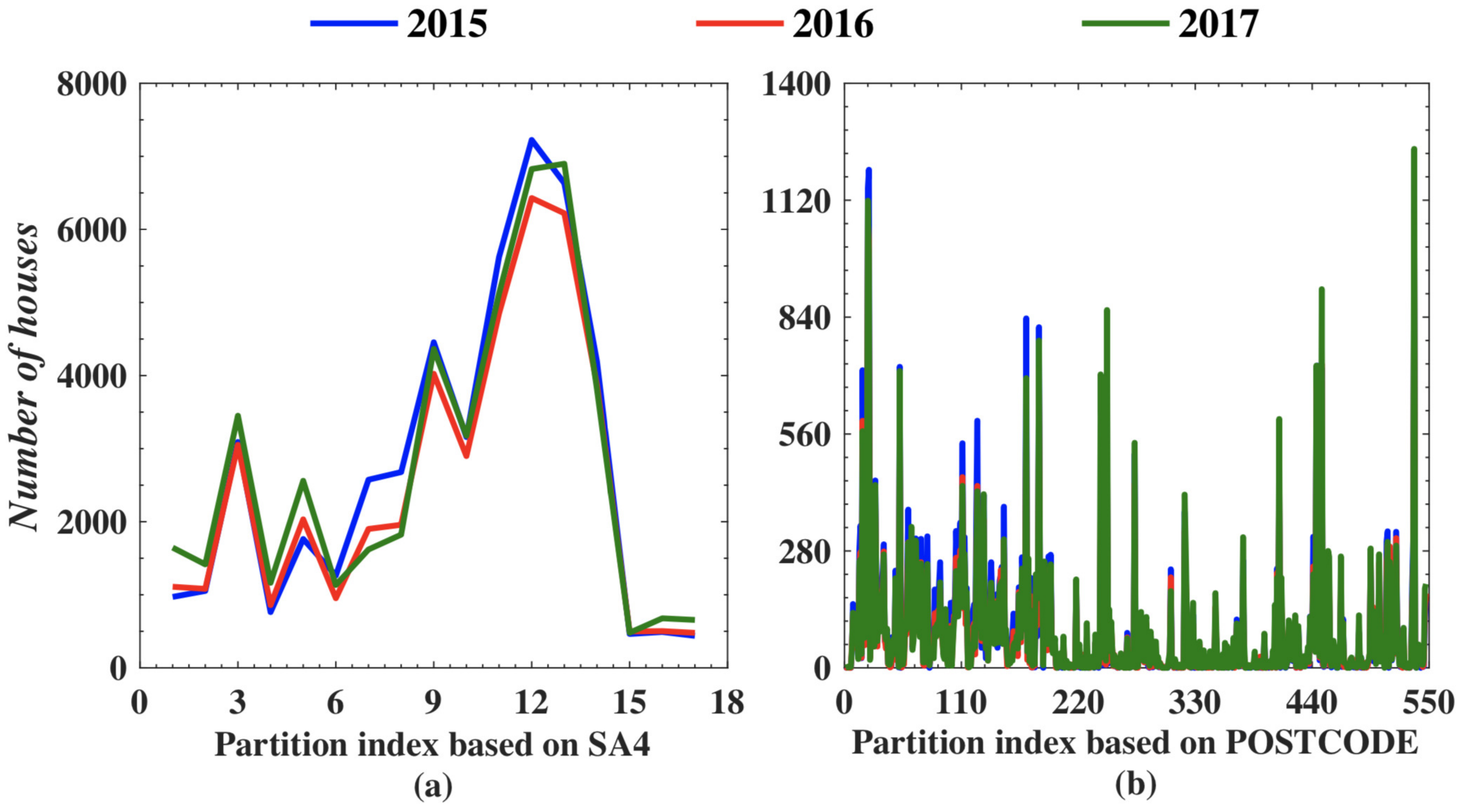}
\caption{Number of samples per partition based on {\SAfourID} and {\POSTCODE}, respectively. The number of samples in the partitions varies greatly. As the split is further refined, the number of samples is generally small.}
\label{fig:sect32}
\end{figure}

The first reason is that no matter which splitting strategy is adopted, it is difficult to ensure that the number of samples allocated to the generated partitions is optimal. We choose partitions based on {\SAfourID} and {\POSTCODE} as two cases, respectively. Figure~\ref{fig:sect32} shows the number of samples for each partition during the three years. We can find that the number of samples in the partitions varies greatly. Moreover, with the further refinement of the split, the number of samples in each partition is generally small, especially in areas where the number of original samples is insufficient, the impact of splitting is more pronounced, which reduces the prediction performance (please refer to the empirical results for the STL-based approaches in Tables~\ref{tab:sect51} and~\ref{tab:sect52} in the Experiments section).

Another important reason is that independent modeling ignores the relatedness between partitions. In fact, partitions are not completely independent, and there are many explicit or implicit connections between them. For example, two independent partitions in geographical space may belong to the same school district. As a result, when analyzing the impact of the school on house price, the two partitions should be merged rather than separated. In order to better illustrate our intuition, we count the number of features that belong to multiple partitions based on {\SAfourID} and {\POSTCODE}, respectively. The results are summarized in Table~\ref{tab:sect34}. The phenomenon that features belong to multiple partitions is very common and becomes more apparent with further splitting. It also indicates the importance of preserving the relatedness between partitions for modeling the problem of house price prediction.

\begin{table}[htbp]
\small\centering
\begin{tabular}{| c | c | c | c | c | c | c |}
\hline
\multirow{2}{*}{Category}
&\multicolumn{3}{c |}{\SAfourID}	&\multicolumn{3}{c |}{\POSTCODE}\\
\cline{2-7}
&=1	&=2	&$\geq$3	&=1	&=2	&$\geq$3\\
\hline
PRIMARY	&\multirow{2}{*}{46}	&\multirow{2}{*}{50}	&\multirow{2}{*}{225}		&\multirow{2}{*}{34}	&\multirow{2}{*}{30}	&\multirow{2}{*}{257}	\\
SCHOOL	&&&&&&\\
SECONDARY	&\multirow{2}{*}{40}	&\multirow{2}{*}{32}	&\multirow{2}{*}{148}		&\multirow{2}{*}{24}	&\multirow{2}{*}{23}	&\multirow{2}{*}{173}	\\
SCHOOL	&&&&&&\\
\hline
{\STATION}	&123		&70	&25	&11	&33	&174	\\
\hline
{\SHOP}	&289		&64	&17	&84	&84	&202	\\
{\HOSPITAL}	&328		&83	&13	&94	&87		&243	\\
{\GP}	&2133	&112	&7	&1386	&633		&233	\\
{\MARKET}	&361	&62	&12	&65	&101		&269	\\
\hline
\end{tabular}
\caption{Number of features that belong to only one partition (=1), two partitions (=2), and at least three partitions (${\geq}$3) based on {\SAfourID} and {\POSTCODE}, respectively.}
\label{tab:sect34}
\end{table}

Based on the above data analysis, we conclude the two challenges that affect the prediction performance of existing approaches from the data view. Meanwhile, considering that the inherent relatedness of the partitions due to the consistency of the house features, we cast the house price prediction problem into an MTL problem. By using the framework of MTL, we can not only reflect the relatedness between partitions well, but also solve the dilemma of insufficient samples in some partitions.

\section{MTL-based House Price Prediction}\label{sect4} 
In this section, we first describe our house price prediction problem and provide preliminaries about MTL. Then we formulate the problem of MTL for house price prediction. To facilitate our illustration, the notations used throughout this paper are presented in Table~\ref{tab:sect41}.

\begin{table}[h!]
\small\centering
\begin{tabular}{| c | p{5.5cm} |}
\hline
Notations	&Explanations\\
\hline
${\pp}$, ${\PP}$	&a task (partition), all tasks (partitions)\\
${\ttt}$, ${\taut}$	&a time interval, a timestamp\\
${\hh}$	&prediction horizon\\
${\mm_{\ttt}^{\pp}}$	&number of samples for task ${\pp}$ in a time interval ${\ttt}$\\
${\mm_{\kk, \ttt}^{\pp}}$	&number of samples for task ${\pp}$ in ${\kk}$ time intervals\\
${\DD}$	&number of features\\
${\xx_{\pp}}$, ${\yy_{\pp}}$	&training input and output for task ${\pp}$\\
${\ww_{\pp}}$, ${\WW}$		&feature weight parameter for task ${\pp}$, weight matrix for all tasks\\
${\LL}, {\OmegaO}$	&empirical loss and regularization error\\
${\rr_{\pp,\qq}}$	 &Ratio of average house price for two tasks ${\pp}$ and ${\qq}$\\
\hline
\end{tabular}
\caption{Notations and explanations.}
\label{tab:sect41}
\end{table}

\subsection{Problem description and preliminaries}
Consider that the house data contains ${\PP}$ partitions and in each time interval ${\ttt}$ (e.g., month, quarter), each partition ${\pp}$ has ${\mm_{\ttt}^{\pp}}$ ${\in}$ ${\mathbb{Z}}$ house transaction records. Therefore, given a timestamp ${\taut}$, the objective of our house price prediction is to predict the price of houses that appear in each partition from ${\taut}$ to ${\taut} + {\hh}$ based on the historical transaction data collected before ${\taut}$, where ${\hh}$ is a specific prediction horizon. It is not difficult to find that multiple timestamps refer to multiple predictions.

In this paper, we comprehend each house price prediction as a problem that is jointly learned by multiple tasks. There are two key points in the formulation of such an MTL: defining tasks and characterizing the relatedness among tasks. In term of the first point, it is usually determined by the information in the data used and the specific application scenario. For example, in the widely used case of predicting student performance in schools, the 139 schools involved were defined as 139 tasks. In term of the second point, methods with regularization terms are generally used to formulate the relatedness among tasks, and different regularization terms represent different ways of formulation. For example, the $l_{2,1}$-norm means that all tasks share a common set of representations.

Here, we define the partitions contained in the house data as tasks, and each partition is aligned with a task. Thus, the two key points in MTL-based house price prediction become how to construct the ${\PP}$ tasks and what methods are used to model the relatedness among these ${\PP}$ tasks. Next, we will introduce our strategies for each of these two issues.

\subsection{Task definition}
In the literature~\cite{zhou2011multi, zhao2015multi, liu2016urban}, tasks are usually uniquely identified and given directly, thus the impact of task definitions on performance has also not received sufficient attention. However, as an essential element of using MTL, there are various ways to define tasks, even in the same application scenario. For example, in the case of predicting student performance in schools as described above, in addition to defining each school as a task, we can also group adjacent schools into one task. Obviously, this change in task definition will affect the subsequent steps of the MTL formulation. Therefore, for our house price prediction problem, we propose two categories of strategies for task definition, and explore the impact of different task definitions on prediction performance.

\subsubsection{Defining tasks based on one single profile}
Existing STL-based house price prediction approaches~\cite{bourassa2010predicting, kuntz2014geostatistical, Montero2018} can be grouped into this category, usually from the perspective of geographical factors. For example, we can define the area of a postal code as one task. One reason for this definition is that geographical factors are the most intuitive expression of house price differences, so task definition along this line is the most common one. Another reason is that the house data used by the above studies includes limited features that affect house price, making it difficult to find more ways to define tasks. Considering that our data set contains a (much richer) variety of house features, we conduct a wide range of task definition attempts and select the following four cases as representatives of this category.

In the house profile, we use the statistical area levels to split the house data. The four area levels indicate four splitting strategies. One partition in each level is defined as a task in the corresponding level. For example, there are 17 partitions at the {\SAfourID} level, so we can get 17 tasks at this level. Similarly, we also consider task definition based on postal code, where one postal code partition corresponds to a task.

In the education profile, we employ the concept of school districts to split the house data, and each school district serves as one task. The primary and secondary school districts lead to two splitting strategies. In addition, we note that the attention to the school district is closely related to the ranking of the school. Therefore, we mainly focus on the school districts of top schools.

In the transportation profile, there seems to be no obvious perspective to define the tasks compared to the previous two feature profiles. Considering that distance/time is an important criterion for measuring the situation of transportation, we define each train station as the centroid of a task and determine the scope of the task by specifying the distance/time threshold to the train station. Therefore, houses with the same train station and the distance/time to the train station (that do not exceed a pre-specified threshold)  belong to the same task.

In the facility profile, we group houses according to the similarity of the facilities. Specifically, there are four types of facilities in our data set, so we give four criteria for measuring similarity, namely {\SharedOne} (share one type), {\SharedTwo} (share two types), {\SharedThree} (share three types), and {\SharedFour} (share four types). Thus, given a criterion and the names of the facilities, such as {\SharedOne}, market, we group houses that have the same market into one task, {\SharedTwo}, shop and hospital, we group houses that have the same shop and hospital into one task.

\subsubsection{Defining tasks based on multiple profiles}
The above task definitions extract one feature profile at a time as a guideline. Such definitions not only constrain the differences in the specific feature profiles of the houses in each task, but also ensure the relatedness among tasks in terms of these profiles. However, the relatedness among tasks that depend on one feature profile are relatively weak. By introducing more feature profiles as a guideline to defining tasks, the relatedness among the resulting tasks can be strengthened, but with the refinement of the definition, the number of houses in each task may be insufficient meanwhile.

In order to guarantee sufficient number of houses necessary for each task and to enhance the relatedness between tasks, we consider six cases by combining any two of the above four task definitions, where each partition obtained corresponds to a task:  
\begin{enumerate}
	\item statistical regions and school districts;
	\item statistical regions and transportation areas; 
	\item statistical regions and neighbor facilities; 
	\item school districts and transportation areas; 
	\item school districts and neighbor facilities; 
	\item transportation areas and neighbor facilities.
\end{enumerate}

\subsection{The MTL model}
In this paper, we regard the MTL-based house price prediction problem as a multi-task regression problem. Given a timestamp ${\taut}$, we extract the transaction records in the previous ${\kk}$ time intervals to construct the training input ${\xx_{\pp}}$ ${\in}$ ${\mathbb{R}^{\mm_{\kk, \ttt}^{\pp}{\times}{\DD}}}$ and output ${\yy_{\pp}}$ ${\in}$ ${\mathbb{R}^{\mm_{\kk, \ttt}^{\pp}{\times}{1}}}$ for each task ${\pp}$. 
Here, ${\mm_{\kk, \ttt}^{\pp}}$ is the number of transaction records in ${\kk}$ time intervals, ${\DD}$ is the number of house features, and ${\yy_{\pp}}$ includes the actual house price. Thus, for each task, we want to infer a linear function ${f_{\pp}}$ where ${f_{\pp}(\xx_{\pp})} = {\xx_{\pp}\ww_{\pp}}$ and ${\ww_{\pp}}$ ${\in}$ ${\mathbb{R}^{\DD{\times}{1}}}$. Let us denote ${\WW = \{\ww_{1}, \ww_{2}, ..., \ww_{\PP}\}}$ ${\in}$ ${\mathbb{R}^{\DD{\times}\PP}}$ as the weight matrix over ${\PP}$ tasks. One typical MTL model for estimating $\WW$ is to minimize the following objective function:
\begin{equation}
\arg\min_{\WW} \LL(\WW) + \OmegaO(\WW),
\label{equ:sect41}
\end{equation}
where ${\LL(\WW)} = {{\sum_{\pp=1}^{\PP}}||\xx_{\pp}\ww_{\pp} - \yy_{\pp}||_{\FF}^{2}}$,  and ${\OmegaO(\WW)}$ is the regularization term that controls the common information shared among tasks.

There can be various choices of regularization terms to fit the above objective function, and the specific choice is based on the identification of the relatedness among the defined tasks. In our house price prediction problem, it is too strict or even unrealistic to use only one type of regularization term because there are various task definitions. Moreover, the purpose of this paper is to study the application of MTL for the house price prediction problem, rather than designing a sophisticated MTL-based method to fit all task definitions. Therefore, we choose three different regularization terms to model the relatedness between tasks, and thus investigate the impact of different MTL-based methods on prediction performance.

The first way is to constrain the models of all tasks to be close to each other. The $l_1$-norm regularization is widely used because it can reduce model complexity and feature learning by introducing sparsity into the model, a common simplification of $l_1$-norm in MTL is that the parameter controlling the sparsity is shared among all tasks. Then the objective function can be defined as:
\begin{equation}
\arg\min_{\WW}{\sum_{\pp=1}^{\PP}}||\xx_{\pp}\ww_{\pp} - \yy_{\pp}||_{\FF}^{2} + \thetaT_{1}||\WW||_{1},
\label{equ:sect42}
\end{equation}
where ${\thetaT_{1}}$ is the parameter that controls sparsity.

The second way is to assume all tasks share a common yet latent representation, such as a common set of features, a common subspace. This motivates the group sparsity, the $l_{2,1}$-norm regularization is usually used to implement this assumption. The objective function can be expressed as:
\begin{equation}
\arg\min_{\WW}{\sum_{\pp=1}^{\PP}}||\xx_{\pp}\ww_{\pp} - \yy_{\pp}||_{\FF}^{2} + \thetaT_{1}||\WW||_{2,1},
\label{equ:sect43}
\end{equation}
where ${\thetaT_{1}}$ is the parameter controlling the group sparsity.

Besides these two most common methods, we also consider ensuring the relatedness between tasks by adding graph regularization. Specifically, the structural relatedness among ${\PP}$ tasks is represented by a graph, each task is defined as a node, and two nodes are connected by a weighted edge. The overall objective function can be described as:
\begin{equation}
\small
\arg\min_{\WW}{\sum_{\pp=1}^{\PP}}||\xx_{\pp}\ww_{\pp} - \yy_{\pp}||_{\FF}^{2} + \thetaT_{1}{\sum_{\pp,\qq=1}^{\PP}{\rr_{\pp,\qq}}||w_\pp - w_\qq||_{\FF}^{2}} + \thetaT_{2}||\WW||_{2,1},
\label{equ:sect44}
\end{equation}
where ${\rr_{\pp,\qq}}$ is the connection strength between nodes (tasks) ${\pp}$ and ${\qq}$, ${\thetaT_{1}}$ and ${\thetaT_{2}}$ are parameters for graph regularization and group sparsity, respectively. 

We define ${\rr_{\pp,\qq}}$ as the ratio of the average house price for the two nodes to measure the structural relatedness between tasks ${\pp}$ and ${\qq}$. Intuitively, the larger the ${\rr_{\pp,\qq}}$, the graph regularization term will force ${\ww_{\pp}}$ to be closer to ${\ww_{\qq}}$. Meanwhile, the closer ${\ww_{\pp}}$ and ${\ww_{\qq}}$ are, the more similar the average house price for these two nodes should be, i.e., ${\rr_{\pp,\qq}}$ tends to 1. Thus, we compute ${\rr_{\pp,\qq}}$ as follows:
\begin{equation}
\rr_{\pp,\qq} = \frac{\min\{average price_{\pp}, average price_{\qq}\}}{\max\{average price_{\pp}, average price_{\qq}\}}.
\label{equ:sect45}
\end{equation}

All the optimization problems above can be solved by using the accelerated gradient descent method~\cite{ruder2016overview}. In this paper, we apply the implementation of accelerated gradient descent method included in the MALSAR~\cite{zhou2011malsar} package to efficiently solve the optimization. 

\section{Experiments}\label{sect5}
In this section, we comprehensively evaluate the performance of our methodology in comparison with five STL-based approaches for house price prediction, and make an in-depth analysis of the impact of task definitions and method selections on our MTL-based house price prediction, and examine the prediction performance of each task individually to demonstrate:
\begin{itemize}
\item{MTL-based methods can significantly outperform the STL-based approaches. (Section~\ref{sect5.3})}

\item{The impact of task definitions on prediction performance far exceeds that of method selections. (Section~\ref{sect5.4})}

\item{The advantages of using MTL to preserve the relatedness among tasks. (Section~\ref{sect5.5})}
\end{itemize}

\subsection{Training and test sets}\label{sect5.1}
The data set studied in this experiment has been detailed in Section~\ref{sect3}. Given the three-year records of sold house data, we first obtain several tasks based on a strategy for defining tasks. Then we evaluate the performance of our methodology by predicting the price of the houses in each task per month, i.e., ${\hh}$ is one month. In each prediction, we create a training set and a test set for each task. On the training set, the data samples are the information tabulation (features, sales price) of the houses that appeared in the previous months of the selected month, i.e., ${\ttt}$ is month. On the test set, we use the data samples in the selected month. Meanwhile, we use the semi-logarithmic form of the sales price to fit the data in the training and test sets.

\begin{figure}[!htbp]
\centering
\includegraphics[width=0.45\textwidth]{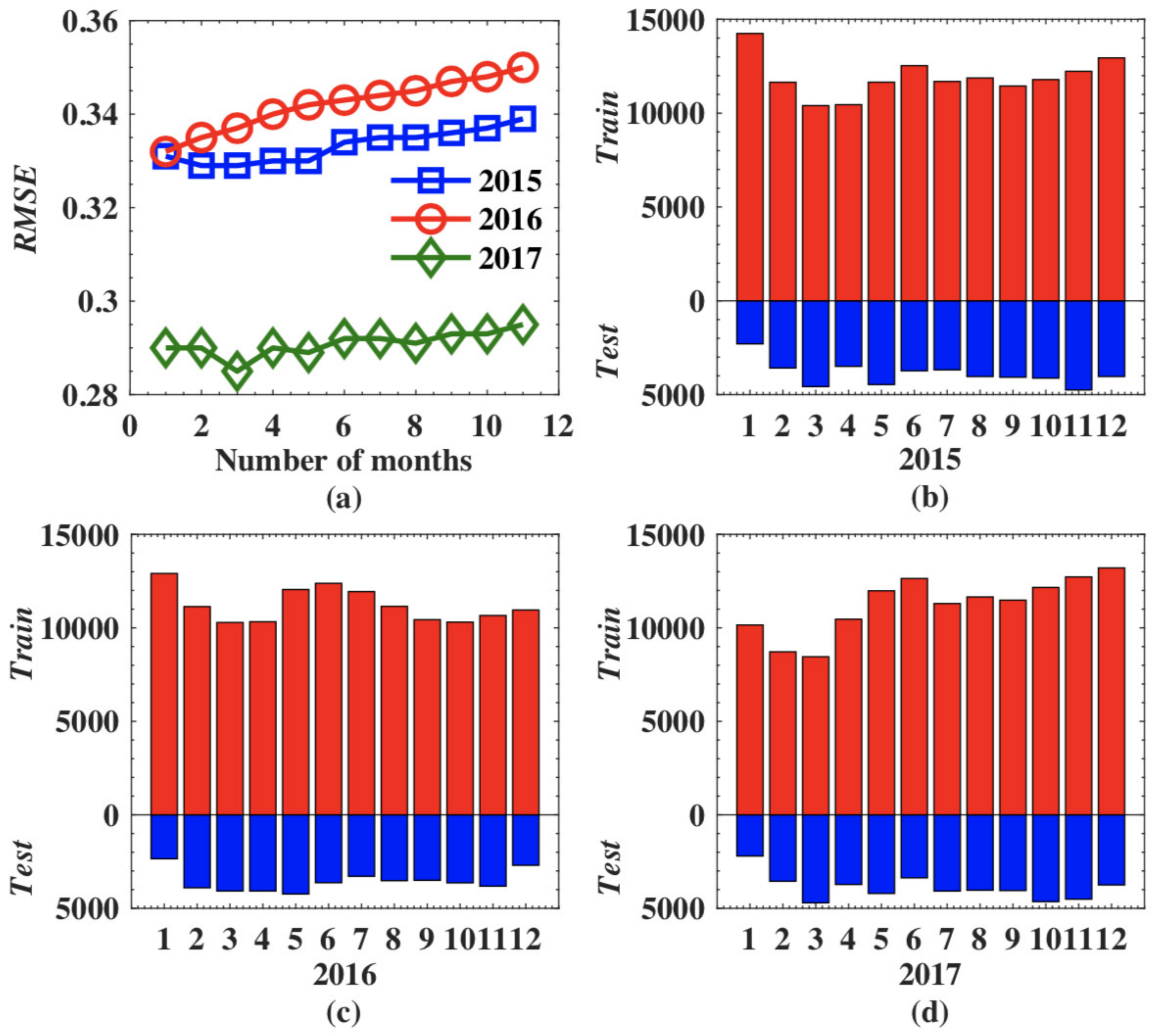}
\caption{Figure (a) shows the impact of the number of previous months used in the training set on the prediction performance for December each year. Figure (b) to Figure (d) show the number of samples in the training and test sets for each prediction.}
\label{fig:sect51}
\end{figure}

In order to analyze the impact of the number of months used for training on prediction, we use the split of the whole data set according to {\SAthreeID} as a case and define the December of each year as the prediction horizon. We first count the number of samples for each task in December and extract those tasks whose number of samples exceeds the first quartile (1/4) of the data distribution as the test set. Then we collect data samples from the previous one month to the previous eleven months as different training sets and fit them using linear regression to predict the price of the houses in December. The prediction performance is shown in Figure~\ref{fig:sect51}a, from which we have two main observations. (1) The prediction error increases with the number of previous months. This trend is not obvious in 2017, as mentioned earlier, the fluctuation of the average monthly house price in 2017 is not significant. (2) The previous three months are the most appropriate one, especially in 2017. In 2015 and 2016, although this does not seem to be the best option, considering that the shorter months may lead to contingency in prediction performance, it is necessary to use the data from the previous three months as training data. Therefore, in the following experiments, unless stated otherwise, we choose to use the previous three months in the training set, i.e., ${\kk}$ is 3.

In summary, we have a total of 36 predictions, each of which is one month in three years. We supplemented the house data for the last three months of 2014 to ensure that the first three months of 2015 are predictable. The number of samples in the training and test sets for each prediction is shown from Figure~\ref{fig:sect51}b to Figure~\ref{fig:sect51}d.

\subsection{Performance metrics}\label{sect5.2} 
To evaluate the prediction performance of different methods, we employ two categories of evaluation metrics in the experiments. 

The first category is two widely used prediction evaluation measures, Root Mean Squared Error~\cite{willmott2005advantages} (RMSE) which considers the square root of the average of all prediction error, and Mean Absolute Error~\cite{chai2014root} (MAE) which calculates the average of absolute error for each predicted result. For each task, these two measures are defined in the following equations.

\begin{equation}
\begin{split}
RMSE_{\pp} &= \sqrt{\frac{1}{N_{\pp}}{\sum_{i=1}^{N_{\pp}}(y_{i}-\hat{y_{i}})^2}},\\
MAE_{\pp} &= \frac{1}{N_{\pp}}{\sum_{i=1}^{N_{\pp}}|y_{i}-\hat{y_{i}}|},
\end{split}
\end{equation}

{\noindent}where ${y_{i}}$ ($\hat{y_{i}}$) is the actual (estimated) house price, ${N_{\pp}}$ is the number of observations for each task ${\pp}$. It is expected that a better prediction will result in a smaller value for both measures.

Given a task definition, we record the metric values for the methods in all tasks and use the mean as the performance of the methods in a prediction. Similarly, the mean of performance in all predictions is used as the overall performance of the methods under this task definition. To make performance comparison in a statistically sound way, we also use Wilcoxon's rank sum test at the significant level of 0.05~\cite{wilcoxon1945individual}.

The second category is the Win-Loss-Draw records~\cite{webb2000multiboosting}. It is a comparative descriptive statistic. The three values, respectively, are the number of data sets for which method ${m_1}$ obtained better, worse, or equal performance outcomes than method ${m_2}$ on a given measure. These summaries compare the performance of the two methods on different data sets and indicate a systematic underlying advantage to one of the methods.

\subsection{Evaluation on overall prediction performance}\label{sect5.3}
We compare our methodology with the following five baseline approaches: 

{\noindent}(1) Lasso Regression~\cite{tibshirani1996regression} ({\Lasso}) is a linear model with $l_{1}$-norm regularization. It tends to prefer solutions that use a small number of features to improve the prediction accuracy and interpretability of the model. Increasing the penalty parameter in {\Lasso} will produce more zeros in the feature coefficients.

{\noindent}(2) Ridge Regression~\cite{hoerl1970ridge} ({\Ridge}) is also a linear model, which gives more penalties by adding $l_{2}$-norm regularization. Although such a formulation loses some information and reduces the accuracy of the fit, the regression coefficients obtained are more realistic and reliable. The penalty parameter in {\Ridge} controls the extent of the loss.

{\noindent}(3) Support Vector Regression~\cite{basak2007support} ({\SVR}) is the natural extension of large margin kernel methods of support vector machine used for classification to regression analysis. It seeks to minimize an upper bound of the generalization error instead of the empirical error. The choice of kernel function, kernel coefficient, and penalty parameter determines the prediction performance of {\SVR}.

{\noindent}(4) AdaBoost~\cite{solomatine2004adaboost} is an algorithm to improve performance by using an ensemble of weak learners to create a strong learner. The output of the weak learners is combined into a weighted sum that represents the final output of the boosted learner. There are many variants of the AdaBoost algorithm depending on the choice of weak learners. Here, we use AdaBoost.R2 ({\AdaRTwo}).

{\noindent}(5) Random Forest~\cite{liaw2002classification} ({\RF}) also belongs to the category of ensemble learning algorithms. Each base learner has the same weight of influence, reducing overfitting by combining them together. Decision tree is often used as the base learner of the ensemble. In this study, the base learner is actually a regression tree. The number of trees in the ensemble is considered to be the design parameter of {\RF}.

We use the Python scikit-learn library~\cite{scikit-learn} to implement the above approaches. There are various parameters and options in each approach. Specifically, we set the kernel function of {\SVR} to a Gaussian function, and the weak learner in {\AdaRTwo} is defined as a decision tree regressor. As for the remaining parameters of each approach, we use 5-fold cross-validation to determine.

Tables~\ref{tab:sect51} and~\ref{tab:sect52} present the values of RMSE and MAE for all tested approaches under different task definitions, and the last row shows the mean of the RMSE and MAE for these approaches. We use the best performing approach under each task definition as a benchmark and compare the performance of the approaches by Wilcoxon's rank sum test, from which we have three main observations. (1) The RMSE and MAE values for all approaches fluctuate insignificantly. This indicates that the performance of these approaches is stable. (2) The first three MTL-based methods outperform all the five baseline approaches. It indicates that the tasks are not independent and capturing their relatedness can improve the learning performance. (3) In the comparison of MTL-based methods, the impact of method changes on overall performance is limited. Although {\MTLGRAPH} enhances the relatedness between tasks through graph regularization, which may help improve performance, we can see from the experimental results that this improvement is not significant.

\subsection{Performance evaluation on different task definitions}\label{sect5.4}
We try two categories of task definitions, i.e., single-profile one and multiple-profile one. For each task definition, we evaluated three MTL-based methods to capture the relatedness between tasks in house price prediction. Given an MTL method, we choose the task definition that makes it perform the best as a benchmark, and compare its performance under various task definitions by Wilcoxon's rank sum test. 

\subsubsection{Task definitions based on one single profile}
The results are summarized in Table~\ref{tab:sect53}. It can be clearly seen that no matter which method is involved, the rank sum test results between its performance under various task definitions and its best performance are similar. For {\MTLONE}, compared to the task definition with the best performance, there are significantly different task definitions that exist in all four single profiles, and are mainly concentrated in the education profile. For {\MTLTWONE}, the experimental phenomenon is very similar to {\MTLONE}. For {\MTLGRAPH}, although the differences between the education profile and the best scenario are still evident, the performance of other profiles has improved.

Comparing these experimental results, we can find: (1) Task definitions in the house price prediction can be diverse. Traditional 

\begin{table*}[h!]
\small\centering
\begin{tabular}
{p{1.8cm} | p{1.8cm}<{\centering} | c | c | c | c | c | c | c | c | c}
\hline
Category	&\multicolumn{2}{c |}{Task definition strategies}	&{\MTLONE}	&{\MTLTWONE}	&{\MTLGRAPH}	&{\Lasso}	&{\Ridge}	&{\SVR}	&{\AdaRTwo}	&{\RF}\\
\hline
\multirow{4}{*}{House}
&\multicolumn{2}{c |}{{\SAfourID}}	&0.219	&0.226	&\textbf{0.192}*	&0.268	&0.260	&0.338	&0.257	&0.244\\
&\multicolumn{2}{c |}{{\SAthreeID}}	&\textbf{0.191}	&\textbf{0.191}	&\textbf{0.189}*	&0.306	&0.280	&0.278	&0.272	&0.225\\
&\multicolumn{2}{c |}{{\SAtwoID}}	&\textbf{0.203}	&\textbf{0.207}	&\textbf{0.190}*	&0.350	&0.310	&0.273	&0.303	&0.223\\
&\multicolumn{2}{c |}{{\POSTCODE}}	&\textbf{0.203}	&\textbf{0.206}	&\textbf{0.191}*	&0.383	&0.366	&0.244	&0.365	&0.214\\
\hline
\multirow{10}{*}{Education}
&\multicolumn{2}{c |}{{\PRIRANK} [1, 10]}	&0.244	&\textbf{0.228}	&\textbf{0.209}*	&0.361	&0.326	&0.325	&0.318	&0.255\\
&\multicolumn{2}{c |}{{\PRIRANK} [1, 20]}	&0.259	&\textbf{0.244}	&\textbf{0.219}*	&0.345	&0.296	&0.351	&0.299	&0.281\\
&\multicolumn{2}{c |}{{\PRIRANK} [1, 30]}	&0.267	&0.253	&\textbf{0.207}*	&0.347	&0.296	&0.372	&0.295	&0.260\\
&\multicolumn{2}{c |}{{\PRIRANK} [1, 40]}	&0.247	&\textbf{0.239}	&\textbf{0.202}*	&0.345	&0.302	&0.364	&0.297	&0.263\\
&\multicolumn{2}{c |}{{\PRIRANK} [1, 50]}	&0.262	&\textbf{0.252}	&\textbf{0.210}*	&0.348	&0.316	&0.353	&0.308	&\textbf{0.257}\\
\cline{2-11}
&\multicolumn{2}{c |}{{\SECRANK} [1, 10]}	&0.263	&\textbf{0.248}	&\textbf{0.229}*	&\textbf{0.252}	&\textbf{0.250}	&0.377	&\textbf{0.249}	&0.274\\
&\multicolumn{2}{c |}{{\SECRANK} [1, 20]}	&\textbf{0.242}	&\textbf{0.229}	&\textbf{0.213}*	&0.255	&0.300	&0.337	&0.285	&0.257\\
&\multicolumn{2}{c |}{{\SECRANK} [1, 30]}	&\textbf{0.235}	&\textbf{0.227}	&\textbf{0.211}*	&0.350	&0.302	&0.340	&0.268	&\textbf{0.247}\\
&\multicolumn{2}{c |}{{\SECRANK} [1, 40]}	&0.262	&0.251	&\textbf{0.218}*	&0.351	&0.307	&0.341	&0.285	&0.277\\
&\multicolumn{2}{c |}{{\SECRANK} [1, 50]}	&0.258	&0.247	&\textbf{0.216}*	&0.354	&0.314	&0.339	&0.290	&0.258\\
\hline
\multirow{5}{*}{Transportation}
&\multicolumn{2}{c |}{{\DISTSTAT} [0, 1000]}		&\textbf{0.213}	&\textbf{0.211}*	&0.221	&0.351	&0.294	&\textbf{0.245}	&0.270	&0.255\\
&\multicolumn{2}{c |}{{\DISTSTAT} [0, 2000]}		&\textbf{0.202}	&\textbf{0.201}*	&\textbf{0.204}	&0.314	&0.313	&0.244	&0.311	&0.230\\
&\multicolumn{2}{c |}{{\DISTSTAT} [0, 3000]}		&\textbf{0.197}*	&\textbf{0.197}	&\textbf{0.198}	&0.323	&0.322	&0.243	&0.322	&\textbf{0.221}\\
&\multicolumn{2}{c |}{{\DISTSTAT} [0, 4000]}		&\textbf{0.194}*	&\textbf{0.195}	&\textbf{0.194}	&0.295	&0.293	&0.250	&0.292	&\textbf{0.217}\\
&\multicolumn{2}{c |}{{\DISTSTAT} [0, 5000]}		&\textbf{0.195}*	&\textbf{0.196}	&\textbf{0.197}	&0.306	&0.303	&0.248	&0.303	&\textbf{0.220}\\
\hline
\multirow{15}{*}{Facility}	
&\multicolumn{2}{c |}{{\SharedOne}\_S ({\SHOP})}	&\textbf{0.199}	&\textbf{0.200}	&\textbf{0.193}*	&0.328	&0.323	&0.262	&\textbf{0.224}	&\textbf{0.221}\\
&\multicolumn{2}{c |}{{\SharedOne}\_H ({\HOSPITAL})}	&\textbf{0.193}	&\textbf{0.192}*	&\textbf{0.192}	&0.329	&0.324	&0.263	&\textbf{0.224}	&\textbf{0.221}\\
&\multicolumn{2}{c |}{{\SharedOne}\_G ({\GP})}	&\textbf{0.201}	&\textbf{0.201}	&\textbf{0.194}*	&0.371	&0.361	&\textbf{0.229}	&0.260	&\textbf{0.226}\\
&\multicolumn{2}{c |}{{\SharedOne}\_M ({\MARKET})}	&\textbf{0.189}	&\textbf{0.188}*	&0.193	&0.316	&0.314	&0.249	&0.216	&0.222\\
\cline{2-11}
&\multicolumn{2}{c |}{{\SharedTwo}\_S, H}		&\textbf{0.193}	&\textbf{0.192}*	&\textbf{0.194}	&0.350	&0.332	&0.243	&\textbf{0.221}	&\textbf{0.222}\\
&\multicolumn{2}{c |}{{\SharedTwo}\_S, G}	&\textbf{0.186}	&\textbf{0.185}*	&\textbf{0.192}	&0.338	&0.336	&0.223	&0.232	&0.228\\
&\multicolumn{2}{c |}{{\SharedTwo}\_S, M}	&\textbf{0.192}	&\textbf{0.191}*	&\textbf{0.194}	&0.341	&0.359	&0.240	&0.235	&\textbf{0.222}\\
&\multicolumn{2}{c |}{{\SharedTwo}\_H, G}	&\textbf{0.188}	&\textbf{0.187}*	&0.194	&0.328	&0.338	&0.224	&0.234	&0.228\\
&\multicolumn{2}{c |}{{\SharedTwo}\_H, M}	&\textbf{0.190}	&\textbf{0.189}*	&\textbf{0.193}	&0.349	&0.337	&0.237	&0.225	&0.223\\
&\multicolumn{2}{c |}{{\SharedTwo}\_G, M}	&\textbf{0.188}	&\textbf{0.187}*	&\textbf{0.194}	&0.335	&0.337	&0.225	&0.234	&0.231\\
\cline{2-11}
&\multicolumn{2}{c |}{{\SharedThree}\_S, H, G}		&\textbf{0.186}	&\textbf{0.185}*	&\textbf{0.192}	&0.347	&0.332	&0.219	&0.228	&0.230\\
&\multicolumn{2}{c |}{{\SharedThree}\_S, H, M}		&\textbf{0.190}	&\textbf{0.189}*	&\textbf{0.194}	&0.346	&0.364	&0.232	&0.260	&0.227\\
&\multicolumn{2}{c |}{{\SharedThree}\_S, G, M}		&\textbf{0.186}	&\textbf{0.185}*	&\textbf{0.192}	&0.332	&0.332	&0.221	&0.219	&0.230\\
&\multicolumn{2}{c |}{{\SharedThree}\_H, G, M}		&\textbf{0.186}	&\textbf{0.185}*	&0.194	&0.346	&0.334	&0.221	&0.230	&0.230\\
\cline{2-11}
&\multicolumn{2}{c |}{{\SharedFour}\_S, H, G, M}	&\textbf{0.185}	&\textbf{0.184}*	&0.192	&0.351	&0.323	&0.217	&0.240	&0.231\\
\hline
\multirow{2}{1.8cm}{House\\Education}
&\multicolumn{2}{c |}{{\SAthreeID}, {\PRIRANK} [1, 40]}	&\textbf{0.194}*	&\textbf{0.201}	&\textbf{0.194}	&0.476	&0.476	&0.266	&0.476	&\textbf{0.206}\\
&\multicolumn{2}{c |}{{\SAthreeID}, {\SECRANK} [1, 30]}	&\textbf{0.195}	&\textbf{0.205}	&\textbf{0.193}*	&0.443	&0.378	&0.273	&0.377	&\textbf{0.208}\\
\hline
\multirow{2}{1.8cm}{House\\Transportation}
&\multicolumn{2}{c |}{\multirow{2}{*}{{\SAthreeID}, [0, 4000]}}	&\multirow{2}{*}{\textbf{0.190}*}	&\multirow{2}{*}{\textbf{0.192}}	&\multirow{2}{*}{\textbf{0.195}}	&\multirow{2}{*}{0.373}	&\multirow{2}{*}{0.353}	&\multirow{2}{*}{0.240}	&\multirow{2}{*}{0.350}	&\multirow{2}{*}{\textbf{0.196}}\\
&\multicolumn{2}{c |}{}&&&&&&&&\\
\hline
\multirow{3}{1.8cm}{House\\Facility}
&\multicolumn{2}{c |}{{\SAthreeID}, M}	&\textbf{0.190}*	&\textbf{0.190}	&\textbf{0.191}	&0.356	&0.339	&0.242	&0.337	&\textbf{0.194}\\
&\multicolumn{2}{c |}{{\SAthreeID}, S, M}	&\textbf{0.193}	&\textbf{0.190}*	&\textbf{0.195}	&0.452	&0.382	&0.237	&0.378	&\textbf{0.197}\\
&\multicolumn{2}{c |}{{\SAthreeID}, S, H, M}	&\textbf{0.191}	&\textbf{0.188}*	&\textbf{0.194}	&0.544	&0.483	&0.230	&0.478	&0.197\\
\hline
\multirow{2}{1.8cm}{Education\\Transportation}
&\multicolumn{2}{c |}{{\PRIRANK} [1, 40], [0, 4000]}	&\textbf{0.190}*	&\textbf{0.191}	&\textbf{0.195}	&0.587	&0.559	&0.234	&0.454	&\textbf{0.197}\\
&\multicolumn{2}{c |}{{\SECRANK} [1, 30], [0, 4000]}	&\textbf{0.191}*	&\textbf{0.194}	&\textbf{0.197}	&0.562	&0.437	&0.240	&0.431	&\textbf{0.199}\\
\hline
\multirow{6}{1.8cm}{Education\\Facility}
&\multicolumn{2}{c |}{{\PRIRANK} [1, 40], M}	&\textbf{0.191}	&\textbf{0.189}*	&0.195	&0.562	&0.437	&0.240	&0.431	&0.198\\
&\multicolumn{2}{c |}{{\PRIRANK} [1, 40], S, M}	&\textbf{0.193}	&\textbf{0.190}*	&\textbf{0.196}	&0.582	&0.584	&0.234	&0.433	&0.200\\
&\multicolumn{2}{c |}{{\PRIRANK} [1, 40], S, H, M}	&\textbf{0.191}	&\textbf{0.187}*	&0.195	&0.639	&0.621	&0.226	&0.418	&0.200\\
&\multicolumn{2}{c |}{{\SECRANK} [1, 30], M}	&\textbf{0.193}*	&\textbf{0.193}	&\textbf{0.193}	&0.617	&0.614	&0.246	&0.470	&\textbf{0.197}\\
&\multicolumn{2}{c |}{{\SECRANK} [1, 30], S, M}	&\textbf{0.193}	&\textbf{0.191}*	&\textbf{0.195}	&0.614	&0.612	&0.237	&0.479	&\textbf{0.198}\\
&\multicolumn{2}{c |}{{\SECRANK} [1, 30], S, H, M}	&\textbf{0.192}	&\textbf{0.190}*	&\textbf{0.195}	&0.629	&0.617	&0.231	&0.445	&0.200\\
\hline
\multirow{3}{1.8cm}{Transportation\\Facility}
&\multicolumn{2}{c |}{[0, 4000], M}	&\textbf{0.192}	&\textbf{0.188}*	&0.198	&0.444	&0.394	&0.228	&0.388	&0.198\\
&\multicolumn{2}{c |}{[0, 4000], S, M}	&\textbf{0.193}	&\textbf{0.188}*	&0.199	&0.685	&0.516	&0.224	&0.462	&0.200\\
&\multicolumn{2}{c |}{[0, 4000], S, H, M}	&\textbf{0.193}	&\textbf{0.188}*	&0.199	&0.678	&0.661	&0.219	&0.471	&0.203\\
\hline
\multicolumn{3}{c |}{Mean of overall performance}	&0.205	&0.203	&0.200	&0.403	&0.378	&0.263	&0.320	&0.220\\
\hline
\end{tabular}
\caption{Evaluation of RMSE among all tested approaches under two categories of task definitions. The first two columns show the specific task definitions, such as House, SA4 means that one partition at the SA4 level is a task. Education, PSCH\_RANK (PRIMARY SCHOOL RANK) [1, 20] means that the top 20 primary schools, one school district is a task. Similarly, SSCH\_RANK (SECONDARY SCHOOL RANK) [1, 40] means that the top 40 secondary schools, one school district is a task. Transportation, STN\_DIS (DISTANCE TO STATION) [0, 4000] means that one station is a task, and houses within 4, 000 meters belong to each task. Facility, SHARED2\_S, M means that houses with the same shop and market belong to the same task. The last eight columns show the RMSE for all tested approaches. In particular, those in bold and asterisk indicate the benchmark under each task definition, and those in bold-only indicate that the p-value for the rank sum test is greater than 0.05.}
\label{tab:sect51}
\end{table*}

\begin{table*}[h!]
\small\centering
\begin{tabular}
{p{1.8cm} | p{1.8cm}<{\centering} | c | c | c | c | c | c | c | c | c}
\hline
Category	&\multicolumn{2}{c |}{Task definition strategies}	&{\MTLONE}	&{\MTLTWONE}	&{\MTLGRAPH}	&{\Lasso}	&{\Ridge}	&{\SVR}	&{\AdaRTwo}	&{\RF}\\
\hline
\multirow{4}{*}{House}
&\multicolumn{2}{c |}{{\SAfourID}}	&0.169	&0.177	&\textbf{0.148}*	&0.208	&0.203	&0.268	&0.200	&0.190\\
&\multicolumn{2}{c |}{{\SAthreeID}}	&\textbf{0.149}	&\textbf{0.150}	&\textbf{0.147}*	&0.231	&0.215	&0.220	&0.208	&0.175\\
&\multicolumn{2}{c |}{{\SAtwoID}}	&\textbf{0.157}	&\textbf{0.161}	&\textbf{0.147}*	&0.261	&0.236	&0.216	&0.230	&0.173\\
&\multicolumn{2}{c |}{{\POSTCODE}}	&\textbf{0.156}	&\textbf{0.161}	&\textbf{0.148}*	&0.288	&0.278	&0.193	&0.277	&\textbf{0.166}\\
\hline
\multirow{10}{*}{Education}
&\multicolumn{2}{c |}{{\PRIRANK} [1, 10]}	&\textbf{0.178}	&\textbf{0.171}	&\textbf{0.150}*	&0.273	&0.235	&0.257	&0.236	&0.185\\
&\multicolumn{2}{c |}{{\PRIRANK} [1, 20]}	&0.193	&0.187	&\textbf{0.164}*	&0.270	&0.225	&0.278	&0.235	&0.205\\
&\multicolumn{2}{c |}{{\PRIRANK} [1, 30]}	&0.199	&0.194	&\textbf{0.157}*	&0.272	&0.228	&0.295	&0.231	&0.197\\
&\multicolumn{2}{c |}{{\PRIRANK} [1, 40]}	&0.187	&0.184	&\textbf{0.154}*	&0.269	&0.269	&0.289	&0.232	&0.202\\
&\multicolumn{2}{c |}{{\PRIRANK} [1, 50]}	&0.197	&0.192	&\textbf{0.160}*	&0.275	&0.242	&0.281	&0.236	&0.195\\
\cline{2-11}
&\multicolumn{2}{c |}{{\SECRANK} [1, 10]}	&0.197	&\textbf{0.192}	&\textbf{0.178}*	&0.195	&\textbf{0.191}	&0.299	&\textbf{0.190}	&0.205\\
&\multicolumn{2}{c |}{{\SECRANK} [1, 20]}	&\textbf{0.184}	&\textbf{0.178}	&\textbf{0.166}*	&0.194	&0.215	&0.267	&0.207	&0.195\\
&\multicolumn{2}{c |}{{\SECRANK} [1, 30]}	&\textbf{0.181}	&\textbf{0.178}	&\textbf{0.166}*	&0.276	&0.225	&0.277	&0.209	&0.189\\
&\multicolumn{2}{c |}{{\SECRANK} [1, 40]}	&0.198	&0.194	&\textbf{0.167}*	&0.274	&0.230	&0.270	&0.219	&0.206\\
&\multicolumn{2}{c |}{{\SECRANK} [1, 50]}	&0.195	&0.192	&\textbf{0.165}*	&0.278	&0.239	&0.267	&0.226	&0.197\\
\hline
\multirow{5}{*}{Transportation}
&\multicolumn{2}{c |}{{\DISTSTAT} [0, 1000]}	&\textbf{0.167}*	&\textbf{0.167}	&\textbf{0.174}	&0.260	&0.229	&0.197	&0.222	&0.199\\
&\multicolumn{2}{c |}{{\DISTSTAT} [0, 2000]}	&\textbf{0.158}*	&\textbf{0.159}	&\textbf{0.160}	&0.237	&0.238	&0.195	&0.238	&0.180\\
&\multicolumn{2}{c |}{{\DISTSTAT} [0, 3000]}	&\textbf{0.155}*	&\textbf{0.155}	&\textbf{0.155}	&0.239	&0.243	&0.194	&0.244	&\textbf{0.173}\\
&\multicolumn{2}{c |}{{\DISTSTAT} [0, 4000]}	&\textbf{0.153}	&\textbf{0.154}	&\textbf{0.152}*	&0.222	&0.223	&0.197	&0.223	&\textbf{0.170}\\
&\multicolumn{2}{c |}{{\DISTSTAT} [0, 5000]}	&\textbf{0.153}*	&\textbf{0.155}	&\textbf{0.154}	&0.229	&0.230	&0.196	&0.230	&\textbf{0.172}\\
\hline
\multirow{15}{*}{Facility}	
&\multicolumn{2}{c |}{{\SharedOne}\_S ({\SHOP})}	&\textbf{0.154}	&\textbf{0.156}	&\textbf{0.149}*	&0.252	&0.243	&0.208	&\textbf{0.175}	&\textbf{0.172}\\
&\multicolumn{2}{c |}{{\SharedOne}\_H ({\HOSPITAL})}	&\textbf{0.150}	&\textbf{0.151}	&\textbf{0.149}*	&0.252	&0.243	&0.209	&\textbf{0.175}	&\textbf{0.171}\\
&\multicolumn{2}{c |}{{\SharedOne}\_G ({\GP})}	&\textbf{0.155}	&\textbf{0.156}	&\textbf{0.149}*	&0.277	&0.275	&0.184	&0.203	&0.178\\
&\multicolumn{2}{c |}{{\SharedOne}\_M ({\MARKET})}	&\textbf{0.150}*	&\textbf{0.150}	&\textbf{0.152}	&0.230	&0.233	&0.197	&\textbf{0.170}	&\textbf{0.171}\\
\cline{2-11}
&\multicolumn{2}{c |}{{\SharedTwo}\_S, H}		&\textbf{0.151}*	&\textbf{0.152}	&\textbf{0.151}	&0.275	&0.264	&0.195	&\textbf{0.172}	&\textbf{0.174}\\
&\multicolumn{2}{c |}{{\SharedTwo}\_S, G}	&\textbf{0.150}*	&\textbf{0.150}	&\textbf{0.153}	&0.266	&0.265	&0.181	&0.184	&0.181\\
&\multicolumn{2}{c |}{{\SharedTwo}\_S, M}	&\textbf{0.151}*	&\textbf{0.151}	&\textbf{0.151}	&0.270	&0.287	&0.191	&0.188	&\textbf{0.173}\\
&\multicolumn{2}{c |}{{\SharedTwo}\_H, G}	&\textbf{0.151}	&\textbf{0.150}*	&\textbf{0.155}	&0.259	&0.267	&0.182	&0.187	&0.181\\
&\multicolumn{2}{c |}{{\SharedTwo}\_H, M}	&\textbf{0.149}*	&\textbf{0.150}	&\textbf{0.151}	&0.275	&0.268	&0.189	&0.178	&\textbf{0.174}\\
&\multicolumn{2}{c |}{{\SharedTwo}\_G, M}	&\textbf{0.150}*	&\textbf{0.150}	&\textbf{0.155}	&0.265	&0.268	&0.182	&0.187	&0.184\\
\cline{2-11}
&\multicolumn{2}{c |}{{\SharedThree}\_S, H, G}		&\textbf{0.151}*	&\textbf{0.151}	&\textbf{0.155}	&0.272	&0.264	&0.180	&0.180	&0.185\\
&\multicolumn{2}{c |}{{\SharedThree}\_S, H, M}		&\textbf{0.151}	&\textbf{0.151}	&\textbf{0.135}*	&0.271	&0.277	&0.187	&0.207	&0.178\\
&\multicolumn{2}{c |}{{\SharedThree}\_S, G, M}		&\textbf{0.150}*	&\textbf{0.150}	&\textbf{0.155}	&0.264	&0.264	&0.180	&\textbf{0.171}	&0.185\\
&\multicolumn{2}{c |}{{\SharedThree}\_H, G, M}		&\textbf{0.151}	&\textbf{0.150}*	&\textbf{0.156}	&0.272	&0.265	&0.181	&0.181	&0.185\\
\cline{2-11}
&\multicolumn{2}{c |}{{\SharedFour}\_S, H, G, M}	&\textbf{0.151}	&\textbf{0.150}*	&\textbf{0.156}	&0.274	&0.243	&0.178	&0.190	&0.186\\
\hline
\multirow{2}{1.8cm}{House\\Education}
&\multicolumn{2}{c |}{{\SAthreeID}, {\PRIRANK} [1, 40]}	&\textbf{0.151}*	&\textbf{0.158}	&\textbf{0.151}	&0.397	&0.396	&0.212	&0.396	&0.164\\
&\multicolumn{2}{c |}{{\SAthreeID}, {\SECRANK} [1, 30]}	&\textbf{0.151}	&\textbf{0.160}	&\textbf{0.150}*	&0.378	&0.312	&0.217	&0.311	&0.165\\
\hline
\multirow{2}{1.8cm}{House\\Transportation}
&\multicolumn{2}{c |}{\multirow{2}{*}{{\SAthreeID}, [0, 4000]}}	&\multirow{2}{*}{\textbf{0.149}*}	&\multirow{2}{*}{\textbf{0.152}}	&\multirow{2}{*}{\textbf{0.153}}	&\multirow{2}{*}{0.281}	&\multirow{2}{*}{0.269}	&\multirow{2}{*}{0.192}	&\multirow{2}{*}{0.266}	&\multirow{2}{*}{\textbf{0.156}}\\
&\multicolumn{2}{c |}{}&&&&&&&&\\
\hline
\multirow{3}{1.8cm}{House\\Facility}
&\multicolumn{2}{c |}{{\SAthreeID}, M}	&\textbf{0.148}*	&\textbf{0.149}	&\textbf{0.149}	&0.263	&0.255	&0.192	&0.253	&\textbf{0.152}\\
&\multicolumn{2}{c |}{{\SAthreeID}, S, M}	&\textbf{0.151}*	&\textbf{0.151}	&\textbf{0.152}	&0.357	&0.298	&0.189	&0.295	&\textbf{0.156}\\
&\multicolumn{2}{c |}{{\SAthreeID}, S, H, M}	&\textbf{0.152}	&\textbf{0.151}*	&\textbf{0.153}	&0.480	&0.390	&0.185	&0.387	&\textbf{0.157}\\
\hline
\multirow{2}{1.8cm}{Education\\Transportation}
&\multicolumn{2}{c |}{{\PRIRANK} [1, 40], [0, 4000]}	&\textbf{0.151}*	&\textbf{0.152}	&\textbf{0.154}	&0.511	&0.492	&0.187	&0.358	&\textbf{0.157}\\
&\multicolumn{2}{c |}{{\SECRANK} [1, 30], [0, 4000]}	&\textbf{0.151}*	&\textbf{0.154}	&\textbf{0.155}	&0.504	&0.371	&0.191	&0.366	&\textbf{0.158}\\
\hline
\multirow{6}{1.8cm}{Education\\Facility}
&\multicolumn{2}{c |}{{\PRIRANK} [1, 40], M}	&\textbf{0.150}*	&\textbf{0.150}	&\textbf{0.153}	&0.504	&0.371	&0.192	&0.366	&\textbf{0.156}\\
&\multicolumn{2}{c |}{{\PRIRANK} [1, 40], S, M}	&\textbf{0.153}	&\textbf{0.152}*	&\textbf{0.155}	&0.509	&0.508	&0.189	&0.368	&0.159\\
&\multicolumn{2}{c |}{{\PRIRANK} [1, 40], S, H, M}	&\textbf{0.153}	&\textbf{0.151}*	&\textbf{0.155}	&0.541	&0.521	&0.184	&0.337	&0.160\\
&\multicolumn{2}{c |}{{\SECRANK} [1, 30], M}	&\textbf{0.151}*	&\textbf{0.152}	&\textbf{0.151}	&0.518*	&0.515	&0.196	&0.380	&\textbf{0.155}\\
&\multicolumn{2}{c |}{{\SECRANK} [1, 30], S, M}	&\textbf{0.152}*	&\textbf{0.152}	&\textbf{0.154}	&0.514	&0.511	&0.191	&0.387	&\textbf{0.158}\\
&\multicolumn{2}{c |}{{\SECRANK} [1, 30], S, H, M}	&\textbf{0.153}	&\textbf{0.152}*	&\textbf{0.155}	&0.525	&0.518	&0.187	&0.347	&0.159\\
\hline
\multirow{3}{1.8cm}{Transportation\\Facility}
&\multicolumn{2}{c |}{[0, 4000], M}	&\textbf{0.152}	&\textbf{0.150}*	&\textbf{0.156}	&0.345	&0.307	&0.184	&0.302	&\textbf{0.156}\\
&\multicolumn{2}{c |}{[0, 4000], S, M}	&\textbf{0.159}	&\textbf{0.152}*	&\textbf{0.159}	&0.573	&0.418	&0.181	&0.374	&0.160\\
&\multicolumn{2}{c |}{[0, 4000], S, H, M}	&\textbf{0.156}	&\textbf{0.154}*	&\textbf{0.161}	&0.569	&0.556	&0.179	&0.383	&0.164\\
\hline
\multicolumn{3}{c |}{Mean of overall performance}	&0.160	&0.158	&0.155	&0.323	&0.300	&0.210	&0.254	&0.175\\
\hline
\end{tabular}
\caption{Evaluation of MAE among all tested approaches under two categories of task definitions. The first two columns show the specific task definitions, such as House, SA4 means that one partition at the SA4 level is a task. Education, PSCH\_RANK (PRIMARY SCHOOL RANK) [1, 20] means that the top 20 primary schools, one school district is a task. Similarly, SSCH\_RANK (SECONDARY SCHOOL RANK) [1, 40] means that the top 40 secondary schools, one school district is a task. Transportation, STN\_DIS (DISTANCE TO STATION) [0, 4000] means that one station is a task, and houses within 4, 000 meters belong to each task. Facility, SHARED2\_S, M means that houses with the same shop and market belong to the same task. The last eight columns show the MAE for all tested approaches. In particular, those in-bold and asterisk indicate the benchmark under each task definition, and those in bold-only indicate that the p-value for the rank sum test is greater than 0.05.}
\label{tab:sect52}
\end{table*}

\begin{table*}[h!]
\small\centering
\begin{tabular}{ p{1.8cm} | c | c | c | c | c | c | c}
\hline
{\multirow{2}{*}{Category}}		&{\multirow{2}{*}{Task definition strategies}}	&\multicolumn{3}{c |}{RMSE}	&\multicolumn{3}{c}{MAE}\\
\cline{3-8}
	&	&{\MTLONE}	&{\MTLTWONE}	&{\MTLGRAPH}	&{\MTLONE}	&{\MTLTWONE}	&{\MTLGRAPH}\\
\hline
\multirow{4}{*}{House}
&{\SAfourID}	&0.219	&0.226	&\textbf{0.192}	&\textbf{0.169}	&\textbf{0.177}	&\textbf{0.148}\\
&{\SAthreeID}	&\textbf{0.191}	&\textbf{0.191}	&\textbf{0.189}*	&\textbf{0.149}	&\textbf{0.150}	&\textbf{0.147}\\
&{\SAtwoID}	&\textbf{0.203}	&\textbf{0.207}	&\textbf{0.190}	&\textbf{0.157}	&\textbf{0.161}	&\textbf{0.147}\\
&{\POSTCODE}	&\textbf{0.203}	&\textbf{0.206}	&\textbf{0.191}	&\textbf{0.156}	&\textbf{0.161}	&\textbf{0.148}\\
\hline
\multirow{10}{*}{Education}
&{\PRIRANK} [1, 10]	&\textbf{0.244}	&\textbf{0.228}	&\textbf{0.209}	&\textbf{0.178}	&\textbf{0.171}	&\textbf{0.150}\\
&{\PRIRANK} [1, 20]	&0.259	&0.244	&0.219	&0.193	&0.187	&0.164\\
&{\PRIRANK} [1, 30]	&0.267	&0.253	&\textbf{0.207}	&0.199	&0.194	&\textbf{0.157}\\
&{\PRIRANK} [1, 40]	&\textbf{0.247}	&\textbf{0.239}	&\textbf{0.202}	&\textbf{0.187}	&0.184	&\textbf{0.154}\\
&{\PRIRANK} [1, 50]	&0.262	&0.252	&0.210	&0.197	&0.192	&\textbf{0.160}\\
\cline{2-8}
&{\SECRANK} [1, 10]	&0.263	&0.248	&0.229	&0.197	&0.192	&0.178\\
&{\SECRANK} [1, 20]	&\textbf{0.242}	&\textbf{0.229}	&\textbf{0.213}	&\textbf{0.184}	&\textbf{0.178}	&\textbf{0.166}\\
&{\SECRANK} [1, 30]	&\textbf{0.235}	&\textbf{0.227}	&\textbf{0.211}	&\textbf{0.181}	&\textbf{0.178}	&\textbf{0.166}\\
&{\SECRANK} [1, 40]	&0.262	&0.251	&0.218	&0.198	&0.194	&\textbf{0.167}\\
&{\SECRANK} [1, 50]	&0.258	&0.247	&0.216	&0.195	&0.192	&\textbf{0.165}\\
\hline
\multirow{5}{*}{Transportation}
&{\DISTSTAT} [0, 1000]	&0.213	&0.211	&0.221	&\textbf{0.167}	&\textbf{0.167}	&0.174\\
&{\DISTSTAT} [0, 2000]	&\textbf{0.202}	&\textbf{0.201}	&\textbf{0.204}	&\textbf{0.158}	&\textbf{0.159}	&\textbf{0.160}\\
&{\DISTSTAT} [0, 3000]	&\textbf{0.197}	&\textbf{0.197}	&\textbf{0.198}	&\textbf{0.155}	&\textbf{0.155}	&\textbf{0.155}\\
&{\DISTSTAT} [0, 4000]	&\textbf{0.194}	&\textbf{0.195}	&\textbf{0.194}	&\textbf{0.153}	&\textbf{0.154}	&\textbf{0.152}\\
&{\DISTSTAT} [0, 5000]	&\textbf{0.195}	&\textbf{0.196}	&\textbf{0.197}	&\textbf{0.153}	&\textbf{0.155}	&\textbf{0.154}\\
\hline
\multirow{15}{*}{Facility}	
&{\SharedOne}\_S ({\SHOP})	&0.199	&0.200	&\textbf{0.193}	&\textbf{0.154}	&\textbf{0.156}	&\textbf{0.149}\\
&{\SharedOne}\_H ({\HOSPITAL})	&\textbf{0.193}	&\textbf{0.192}	&\textbf{0.192}	&\textbf{0.150}	&\textbf{0.151}	&\textbf{0.149}\\
&{\SharedOne}\_G ({\GP})		&0.201	&0.201	&\textbf{0.194}	&\textbf{0.155}	&\textbf{0.156}	&\textbf{0.149}\\
&{\SharedOne}\_M ({\MARKET})	&\textbf{0.189}	&\textbf{0.188}	&\textbf{0.193}	&\textbf{0.150}	&\textbf{0.150}	&\textbf{0.152}\\
\cline{2-8}
&{\SharedTwo}\_S, H	&\textbf{0.193}	&0.192	&\textbf{0.194}	&\textbf{0.151}	&\textbf{0.152}	&\textbf{0.151}\\
&{\SharedTwo}\_S, G	&\textbf{0.186}	&\textbf{0.185}	&\textbf{0.192}	&\textbf{0.150}	&\textbf{0.150}	&\textbf{0.153}\\
&{\SharedTwo}\_S, M	&\textbf{0.192}	&\textbf{0.191}	&\textbf{0.194}	&\textbf{0.151}	&\textbf{0.151}	&\textbf{0.151}\\
&{\SharedTwo}\_H, G	&\textbf{0.188}	&\textbf{0.187}	&0.194	&\textbf{0.151}	&\textbf{0.150}	&\textbf{0.155}\\
&{\SharedTwo}\_H, M	&\textbf{0.190}	&\textbf{0.189}	&\textbf{0.193}	&\textbf{0.149}	&\textbf{0.150}	&\textbf{0.151}\\
&{\SharedTwo}\_G, M	&\textbf{0.188}	&\textbf{0.187}	&\textbf{0.194}	&\textbf{0.150}	&\textbf{0.150}	&\textbf{0.155}\\
\cline{2-8}
&{\SharedThree}\_S, H, G		&\textbf{0.186}	&\textbf{0.185}	&\textbf{0.192}	&\textbf{0.151}	&\textbf{0.151}	&\textbf{0.155}\\
&{\SharedThree}\_S, H, M		&\textbf{0.190}	&\textbf{0.189}	&\textbf{0.194}	&\textbf{0.151}	&\textbf{0.151}	&\textbf{0.135}*\\
&{\SharedThree}\_S, G, M		&\textbf{0.186}	&\textbf{0.185}	&\textbf{0.192}	&\textbf{0.150}	&\textbf{0.150}	&\textbf{0.155}\\
&{\SharedThree}\_H, G, M	&\textbf{0.186}	&\textbf{0.185}	&\textbf{0.194}	&\textbf{0.151}	&\textbf{0.150}	&\textbf{0.156}\\
\cline{2-8}
&{\SharedFour}\_S, H, G, M	&\textbf{0.185}*	&\textbf{0.184}*	&\textbf{0.192}	&\textbf{0.151}	&\textbf{0.150}	&\textbf{0.156}\\
\hline
\multirow{2}{1.8cm}{House\\Education}
&{\SAthreeID}, {\PRIRANK} [1, 40]	&\textbf{0.194}	&\textbf{0.201}	&\textbf{0.194}	&\textbf{0.151}	&\textbf{0.158}	&\textbf{0.151}\\
&{\SAthreeID}, {\SECRANK} [1, 30]	&\textbf{0.195}	&0.205	&0.193	&\textbf{0.151}	&0.160	&\textbf{0.150}\\
\hline
\multirow{2}{1.8cm}{House\\Transportation}
&\multirow{2}{*}{{\SAthreeID}, [0, 4000]}	&\multirow{2}{*}{\textbf{0.190}}	&\multirow{2}{*}{\textbf{0.192}}	&\multirow{2}{*}{\textbf{0.195}}	&\multirow{2}{*}{\textbf{0.149}}	&\multirow{2}{*}{\textbf{0.152}}	&\multirow{2}{*}{\textbf{0.153}}\\
&&&&&&&\\
\hline
\multirow{3}{1.8cm}{House\\Facility}
&{\SAthreeID}, M	&\textbf{0.190}	&\textbf{0.190}	&\textbf{0.191}	&\textbf{0.148}*	&\textbf{0.149}*	&\textbf{0.149}\\
&{\SAthreeID}, S, M	&\textbf{0.193}	&\textbf{0.190}	&\textbf{0.195}	&\textbf{0.151}	&\textbf{0.151}	&\textbf{0.152}\\
&{\SAthreeID}, S, H, M	&\textbf{0.191}	&\textbf{0.188}	&\textbf{0.194}	&\textbf{0.152}	&\textbf{0.151}	&\textbf{0.153}\\
\hline
\multirow{2}{1.8cm}{Education\\Transportation}
&{\PRIRANK} [1, 40], [0, 4000]	&\textbf{0.190}	&\textbf{0.191}	&\textbf{0.195}	&\textbf{0.151}	&\textbf{0.152}	&\textbf{0.154}\\
&{\SECRANK} [1, 30], [0, 4000]	&\textbf{0.191}	&\textbf{0.194}	&\textbf{0.197}	&\textbf{0.151}	&0.154	&\textbf{0.155}\\
\hline
\multirow{6}{1.8cm}{Education\\Facility}
&{\PRIRANK} [1, 40], M	&\textbf{0.191}	&\textbf{0.189}	&\textbf{0.195}	&\textbf{0.150}	&\textbf{0.150}	&\textbf{0.153}\\
&{\PRIRANK} [1, 40], S, M	&\textbf{0.193}	&\textbf{0.190}	&\textbf{0.196}	&\textbf{0.153}	&\textbf{0.152}	&\textbf{0.155}\\
&{\PRIRANK} [1, 40], S, H, M		&\textbf{0.191}	&\textbf{0.187}	&\textbf{0.195}	&\textbf{0.153}	&\textbf{0.151}	&\textbf{0.155}\\
&{\SECRANK} [1, 30], M	&\textbf{0.193}	&\textbf{0.193}	&\textbf{0.193}	&\textbf{0.151}	&\textbf{0.152}	&\textbf{0.151}\\
&{\SECRANK} [1, 30], S, M	&\textbf{0.193}	&\textbf{0.191}	&\textbf{0.195}	&\textbf{0.152}	&\textbf{0.152}	&\textbf{0.154}\\
&{\SECRANK} [1, 30], S, H, M		&\textbf{0.192}	&\textbf{0.190}	&\textbf{0.195}	&\textbf{0.153}	&\textbf{0.152}	&\textbf{0.155}\\
\hline
\multirow{3}{1.8cm}{Transportation\\Facility}
&[0, 4000], M	&\textbf{0.192}	&\textbf{0.188}	&\textbf{0.198}	&\textbf{0.152}	&\textbf{0.150}	&\textbf{0.156}\\
&[0, 4000], S, M	&\textbf{0.193}	&\textbf{0.188}	&0.199	&0.159	&\textbf{0.152}	&\textbf{0.157}\\
&[0, 4000], S, H, M	&\textbf{0.193}	&\textbf{0.188}	&0.199	&0.156	&\textbf{0.154}	&0.161\\
\hline
\end{tabular}
\caption{Evaluation of RMSE and MAE among three MTL-based methods under two categories of task definitions. The first two columns show the specific task definitions, such as House, Transportation: SA3, [0, 4000] means that the tasks are defined based on these two profiles, and the same SA3 and houses within 4000 meters of the same station belong to the same task. The last six columns show the RMSE and MAE for three MTL-based methods. In particular, those in bold and asterisk indicate the benchmark under each method, and those in bold-only indicate that the p-value for the rank sum test is greater than 0.05.}
\label{tab:sect53}
\end{table*}

{\noindent}prediction approaches are usually considered from geographical factors, but it can be seen from the experimental results that other factors can also ensure good prediction performance, even beyond geographical factors. (2) The optimal MTL-based methods that correspond to the various task definitions are different. Although {\MTLGRAPH} enhances the relatedness between tasks by adding graph regularization, such a setting is too strict in some task definitions, which reduces the prediction performance. For example, in the transportation profile, compared with {\MTLGRAPH}, the performance of {\MTLONE} and {\MTLTWONE} is available. (3) In terms of prediction performance, the choice of task definitions has a greater impact than the choice of MTL-based methods, such as, the performance of the facility profile is superior to the education profile regardless of which MTL-based method is used.

\subsubsection{Task definitions based on multiple profiles}
We choose the following task definitions as representatives of each profile and assemble them in pairs to redefine tasks. The first is {\SAthreeID} in the statistical regions. The second is {\PRIRANK} [1, 40], {\SECRANK} [1, 30] in the primary and secondary school districts, respectively. The third is {\DISTSTAT} [0, 4000] in the transportation areas. The fourth is M ({\MARKET}) in {\SharedOne}, S ({\SHOP}), M in {\SharedTwo}, and S, H ({\HOSPITAL}), M in {\SharedThree} in the neighbor facilities, respectively. The prediction 

\begin{table*}
\small\centering
\begin{tabular}
{c | c | c | c | c | c | c | c | c | c}
\hline
Task definition strategies	&Group	&{\MTLONE}	&{\MTLTWONE}	&{\MTLGRAPH}	&{\Lasso}	&{\Ridge}	&{\SVR}	&{\AdaRTwo}	&{\RF}\\
\hline
\multirow{4}{*}{\SAthreeID}
&(0, 1/4]	&6/11/0	&5/12/0	&\textbf{9/8/0}	&0/17/0	&0/17/0	&1/16/0	&0/17/0	&0/17/0\\
&(1/4, 1/2]		&6/9/1	&4/12/0	&\textbf{9/6/1}	&0/16/0	&0/16/0	&0/16/0	&0/16/0	&0/16/0\\
&(1/2, 3/4]		&\textbf{12/4/0}	&3/13/0	&4/12/0	&0/16/0	&0/16/0	&0/16/0	&0/16/0	&0/16/0\\
&(3/4, 1]	&\textbf{10/4/2}	&0/16/0	&4/10/2	&0/16/0	&0/16/0	&0/16/0	&0/16/0	&0/16/0\\
\hline
\multirow{4}{*}{{\SECRANK} [1, 30]}
&(0, 1/4]	&1/4/0	&0/5/0	&\textbf{3/2/0}	&0/5/0	&0/5/0	&1/4/0	&0/5/0	&0/5/0\\
&(1/4, 1/2]		&3/2/0	&\textbf{4/1/0}	&1/4/0	&0/5/0	&1/4/0	&0/5/0	&1/4/0	&1/4/0\\
&(1/2, 3/4]		&\textbf{2/1/0}	&\textbf{2/1/0}	&0/3/0	&0/3/0	&1/2/0	&0/3/0	&1/2/0	&0/3/0\\
&(3/4, 1]	&3/2/0	&4/1/0	&0/5/0	&0/5/0	&2/3/0	&\textbf{4/1/0}	&2/3/0	&1/4/0\\
\hline
\multirow{4}{*}{{\DISTSTAT} [0, 4000]}
&(0, 1/4]	&\textbf{24/29/1}	&\textbf{24/29/1}	&18/35/1	&1/53/0	&2/52/0	&14/39/1	&1/53/0	&17/36/1\\
&(1/4, 1/2]		&20/34/0	&20/34/0	&\textbf{22/30/2}	&0/54/0	&0/54/0	&6/48/0	&0/54/0	&21/31/2\\
&(1/2, 3/4]		&25/26/3	&25/26/3	&15/38/1	&0/54/0	&0/54/0	&2/52/0	&0/54/0	&\textbf{34/20/0}\\
&(3/4, 1]	&18/36/0	&18/36/0	&\textbf{28/24/2}	&0/54/0	&0/54/0	&1/52/1	&0/54/0	&9/41/4\\
\hline
\multirow{4}{*}{\MARKET}	
&(0, 1/4]	&52/53/2	&\textbf{61/42/4}	&17/85/5	&1/106/0	&2/105/0	&26/80/1	&3/104/0	&47/57/3\\
&(1/4, 1/2]		&39/68/1	&\textbf{47/57/4}	&33/73/2	&0/108/0	&1/107/0	&6/101/1	&1/107/0	&46/60/2\\
&(1/2, 3/4]		&33/71/2	&39/65/2	&\textbf{44/59/3}	&0/106/0	&0/106/0	&2/104/0	&0/106/0	&33/69/4\\
&(3/4, 1]	&29/72/5	&28/76/2	&\textbf{51/48/7}	&0/106/0	&0/106/0	&1/105/0	&0/106/0	&22/78/6\\
\hline
\end{tabular}
\caption{Win/Loss/Draw records obtained by comparing all tested approaches with {\MTLGRAPH} based on four different task definitions. The first two columns show the selected task definitions and the quantile-based groupings, such as {\SAthreeID}, (1/4, 1/2] means that the tasks are defined based on {\SAthreeID}, and the number of samples for the tasks in this group is between the first quartile and the second quartile of the distribution of the number of samples for all tasks. The last eight columns show the Win/Loss/Draw scores for all tested approaches. Those in bold correspond to the best scenarios.}
\label{tab:sect54}
\end{table*}

\begin{figure*}
\centering
\subfloat[{Task definition based on {\SAthreeID}}]{\includegraphics[width=0.45\textwidth]{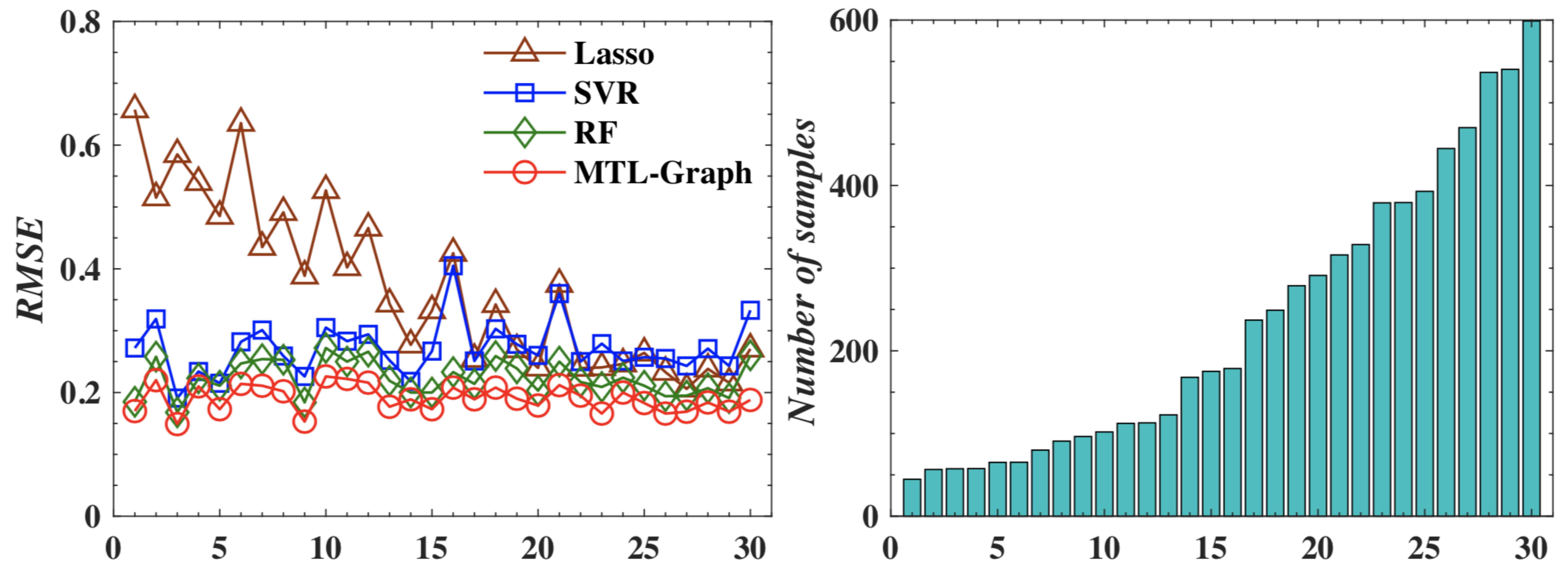}
\label{fig:sect52a}}
\hfil
\subfloat[{Task definition based on {\SECRANK} [1, 30]}]{\includegraphics[width=0.45\textwidth]{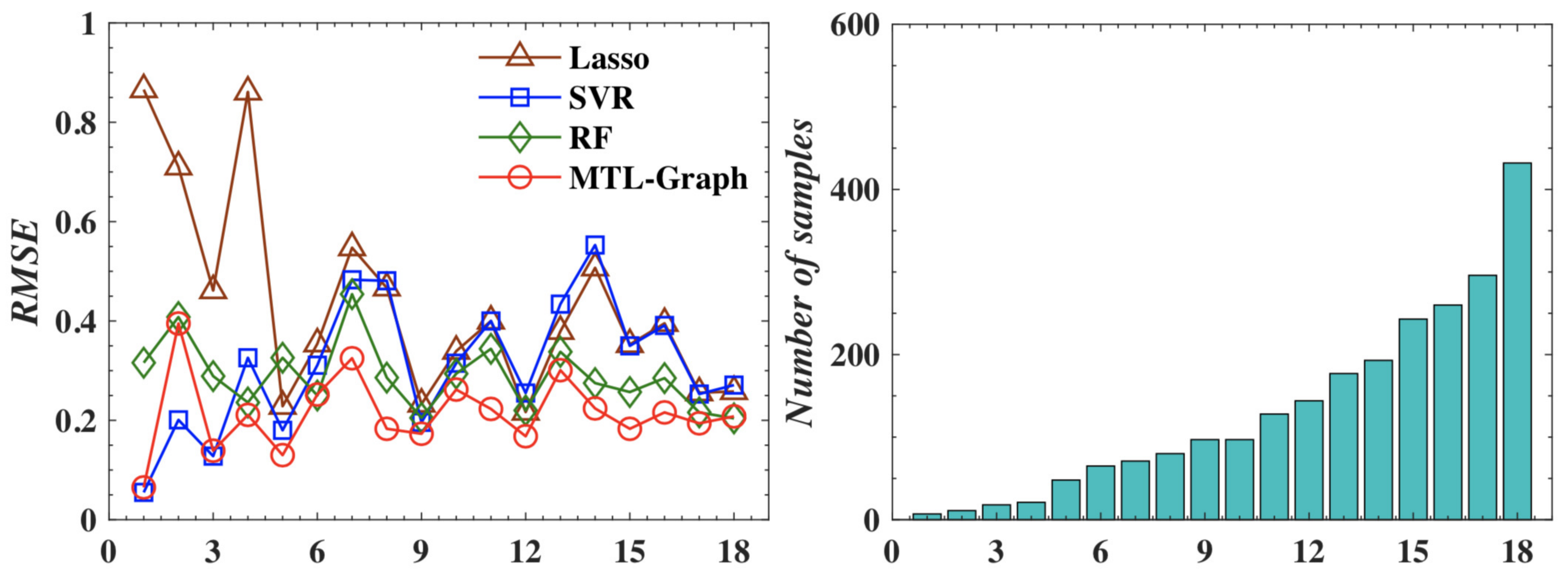}
\label{fig:sect52b}}
\hfil
\subfloat[{Task definition based on {\DISTSTAT} [0, 4000]}]{\includegraphics[width=0.45\textwidth]{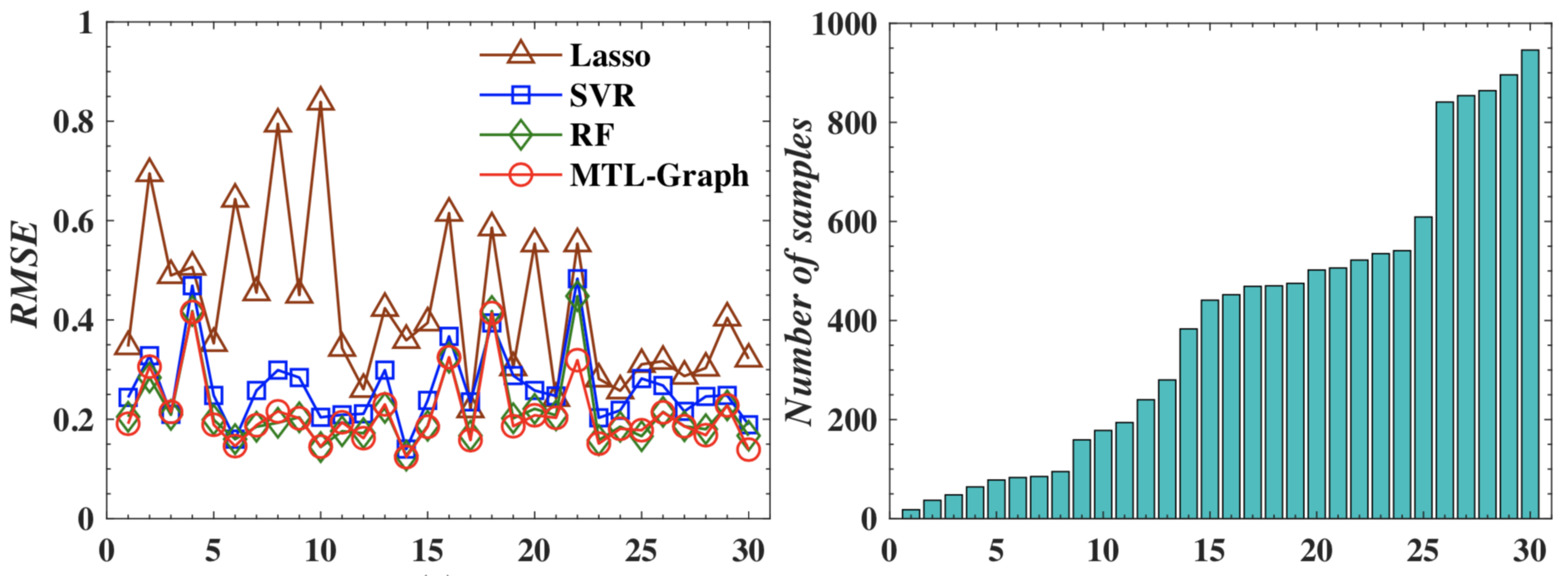}
\label{fig:sect52c}}
\hfil
\subfloat[{Task definition based on {\MARKET}}]{\includegraphics[width=0.45\textwidth]{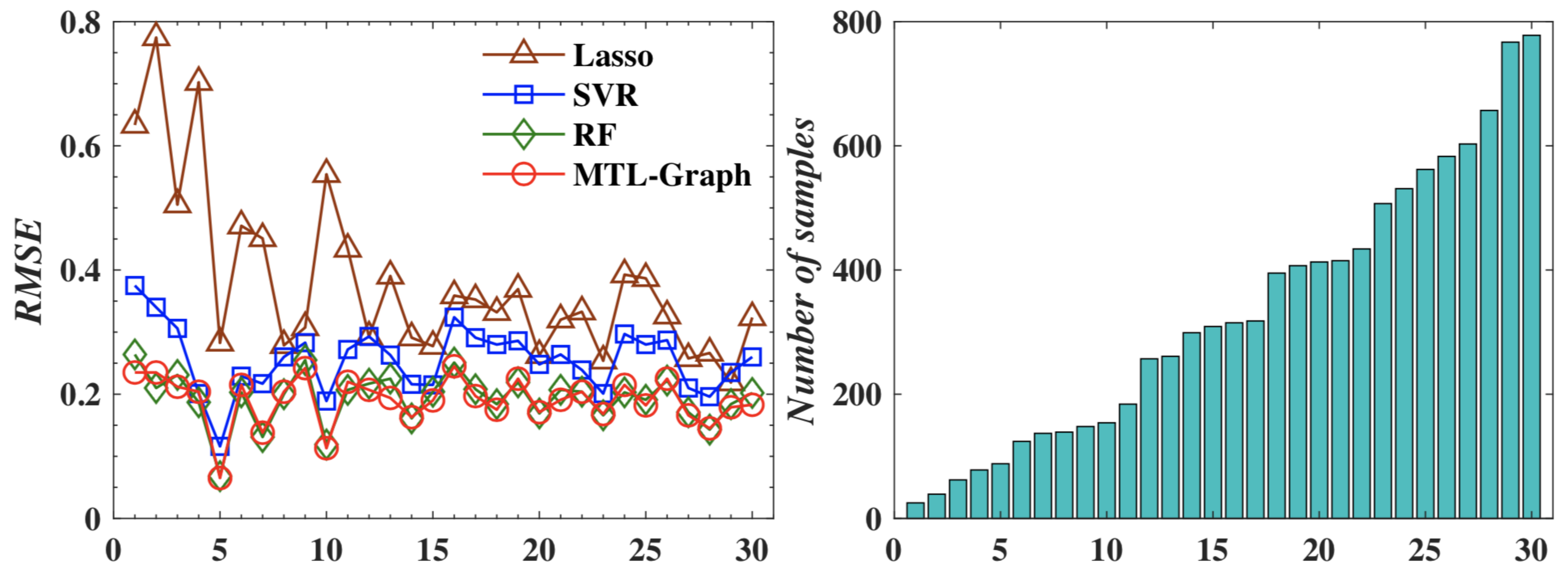}
\label{fig:sect52d}}
\caption{Evaluation of the prediction performance for four methods in each task. The x-axis corresponds to the task indexes under the specified task definition. The y-axis corresponds to the RMSE and the number of samples per task, respectively.}
\label{fig:sect52}
\end{figure*}

{\noindent}performance of MTL-based methods under each of the above task definitions is also summarized in Table~\ref{tab:sect53}.

We have the following observations. (1) The performance difference between various task definitions is not significant. This is because the definition based on multiple profiles is fine-grained, hence the tasks obtained are similar, which makes the difference in performance small. (2) The performance of {\MTLONE} and {\MTLTWONE} is closer to {\MTLGRAPH}, even better in some task definitions. This somehow indicates that optimizing the performance by enhancing the relatedness among tasks is limited when the task definitions are sufficient. (3) Compared with one single profile, the overall performance based on multiple profiles is generally better. This also indicates that the task definitions have a greater impact on prediction performance than that of the MTL-based methods.

\vspace{0.5em}\noindent\textbf{Discussion.} Based on the above analysis, we not only validate the impact of task definitions and method selections on prediction performance, but also demonstrate that the impact of task definitions on prediction performance far exceeds that of method selections. In addition, we note that the MTL-based method using graph regularization usually performs well when the task definitions are not sufficient (definition based on one single profile). However, when the task definitions are very refined (definition based on multiple profiles), the performance of the MTL-based methods using general regularization ($l_{1}$-norm, $l_{2,1}$-norm) is good enough, which deserves further investigations via more data sources.

\subsection{Performance evaluation on each task}\label{sect5.5}
Why MTL-based prediction methods can achieve good performance? We here take a further look into the MTL model by investigating the quality of prediction for each task in a prediction. In particular, we extract the task definitions based on {\SAthreeID}, {\SECRANK} [1, 30], {\DISTSTAT} [0, 4000], and M ({\MARKET}) in {\SharedOne} as four cases, respectively. In each task definition, we first count the distribution of the number of data samples for all tasks, and then divide the tasks into four groups according to the first quartile (1/4), the second quartile (1/2), and the third quartile (3/4). Finally, we choose {\MTLGRAPH} as a benchmark and choose RMSE as the measure to count the Win-Loss-Draw records for all tested approaches in each group. 

The results are summarized in Table~\ref{tab:sect54}. We can find: (1) In groups with fewer samples, i.e., (0, 1/4] and (1/4, 1/2], the performance of MTL-based methods is significantly better than that of STL-based approaches. This confirms the advantages of MTL model. (2) In groups with enough samples, i.e., (1/2, 3/4] and (3/4, 1], the performance of MTL-based methods is still good, even in some task definitions, {\SVR} and {\RF} outperform MTL-based methods. However, STL-based approaches ignore the relatedness between tasks, which result in well-behaved approaches in these groups that do not fit into other groups. Therefore, the overall performance of MTL-based methods is superior to them.

In order to better understand the conclusions of the above quantitative evaluation, we illustrate the performance of {\Lasso}, {\SVR}, {\RF}, and {\MTLGRAPH} for each task based on four task definitions. As shown in Figure~\ref{fig:sect52}, we extract 30 tasks under each task definition and give the number of samples for each task and the prediction performance for four approaches in each task. Note that the results of some SECONDARY SCHOOL cannot be described in Figure~\ref{fig:sect52b} due to the absence of house transactions. Taking Figure~\ref{fig:sect52a} as an example, it can be clearly seen: (1) In tasks with fewer samples, such as task index less than 10, the performance of MTL-based methods is consistently better than that of other methods. (2) The performance gap is narrowed as the number of samples in the tasks increases, such as task index greater than 22, but the performance of MTL-based methods is still acceptable.   

In summary, our MTL-based house price prediction is robust. It guarantees the prediction performance when data samples are sufficient, and when the samples are insufficient, it optimizes the prediction performance by exploiting the relatedness between tasks. As a result, the overall performance is improved.

\section{Conclusions and Future Work}\label{sect6}
In this paper, we carried out a deep research on the application of MTL for the house price prediction problem. In terms of data profiling, we define and capture a fine-grained location profile powered by a diverse range of location data sources. In term of prediction model, there are two key points in the implementation of MTL-based house price prediction: task definitions and method selections. Therefore, we designed two categories of strategies to define tasks based on various house features, and selected three general MTL-based methods with different regularization terms to capture and utilize the relatedness between tasks. By extensive experimental evaluations, we first demonstrated that modeling based on MTL can significantly improve the overall performance of house price prediction. Then we illustrated that the diversity of task definitions is conducive to the MTL formulation for the house price prediction problem. Finally, we revealed that the impact of task definitions on prediction performance far exceeds that of method selections.

In the future, we will extend our methodology to adaptively learn task definitions, and we plan to explore non-linear models rather than only focus on linear models. We will also try the house recommendations based on the outcomes of the price prediction. 

\section*{Acknowledgments}
This work was partially supported by ARC under Grants DP170102726, DP180102050, DP170102231, DP160103595, and the National Natural Science Foundation of China (NSFC) under Grants 61728204, 91646204, and 71571093. Zhifeng Bao is a recipient of Google Faculty Award.

\bibliographystyle{IEEEtran}
\bibliography{paperRefs}

\begin{IEEEbiography}[{\includegraphics[width=1in,height=1.25in,clip,keepaspectratio]{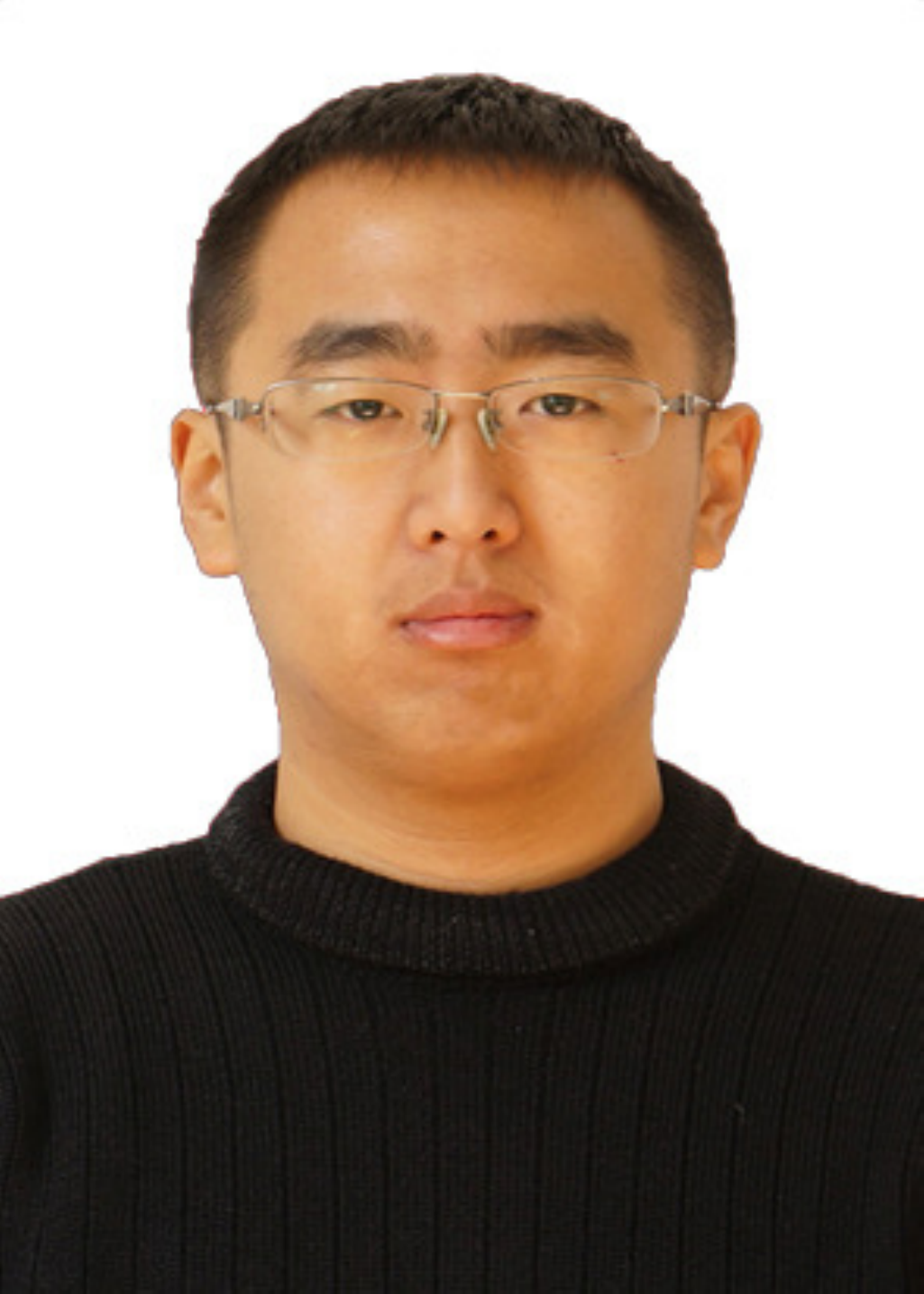}}]
{Guangliang Gao} received the B.E. degree from the Anhui University of Science and Technology, Anhui, China, in 2012. He is currently pursuing the Ph.D. degree with the Nanjing University of Science and Technology, Nanjing, China. He is currently a visiting student at Computer Science and Information Technology, RMIT University, Melbourne, VIC, Australia. His current research interests include data mining and machine learning over social network data and geo-spatial data.
\end{IEEEbiography}\vspace{-1.3cm}

\begin{IEEEbiography}[{\includegraphics[width=1in,height=1.25in,clip,keepaspectratio]{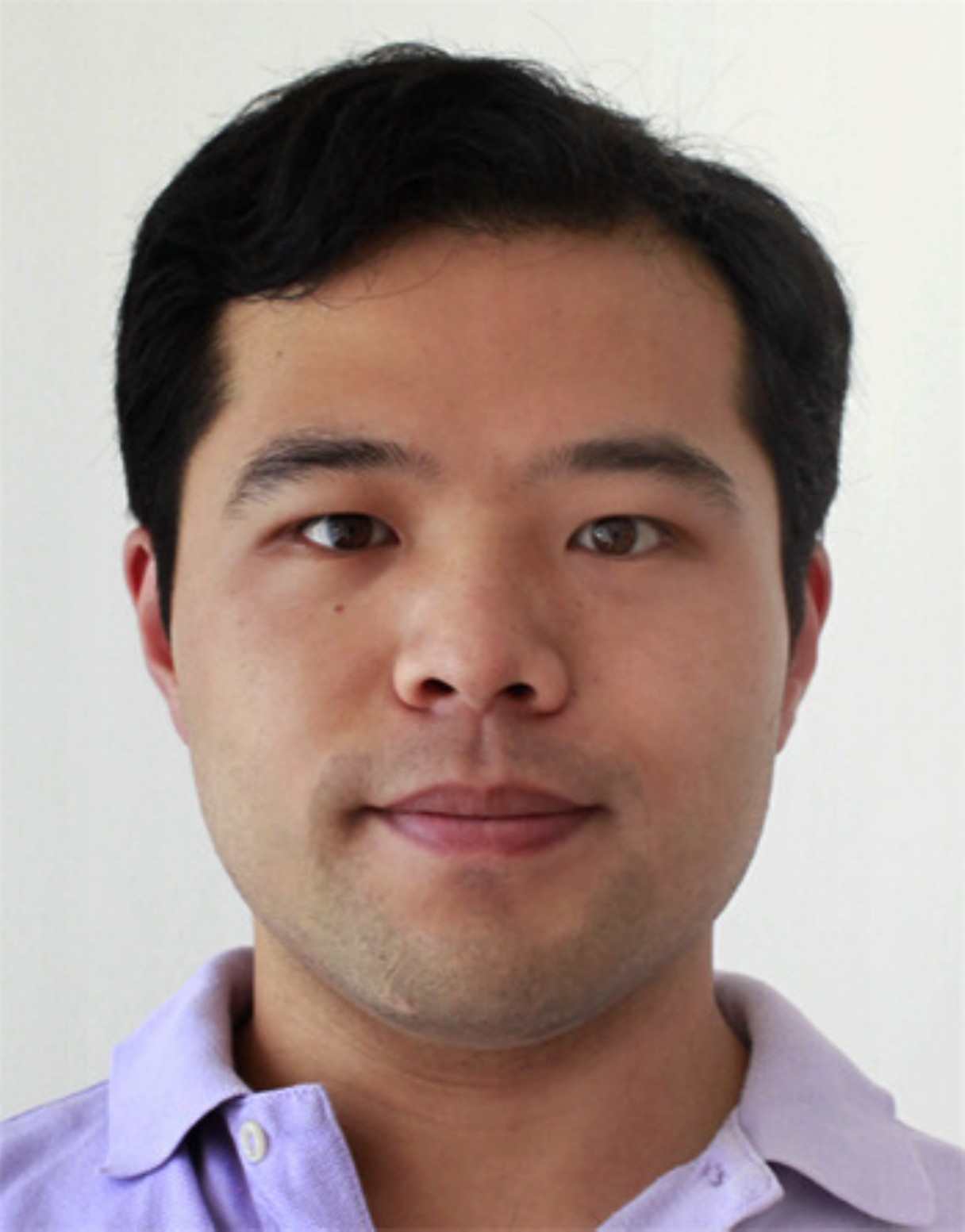}}]
{Zhifeng Bao} received the Ph.D. degree in computer science from the National University of Singapore in 2011 as the winner of the Best PhD Thesis in school of computing. He is currently a senior lecturer with the RMIT University and leads the big data research group at RMIT. He is also an Honorary Fellow with University of Melbourne in Australia. His current research interests include data usability, spatial database, data integration, and data visualization.
\end{IEEEbiography}\vspace{-1.3cm}

\begin{IEEEbiography}[{\includegraphics[width=1in,height=1.25in,clip,keepaspectratio]{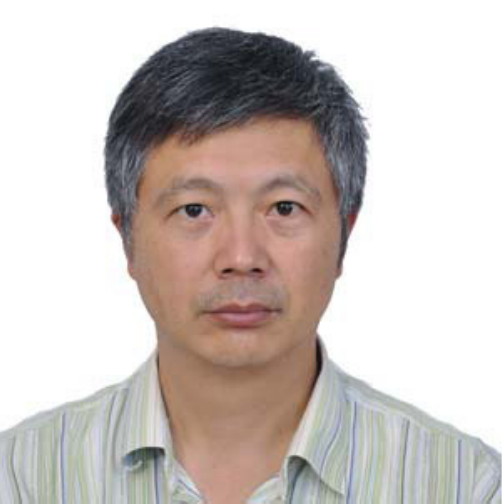}}]
{Jie Cao} received his Ph.D. degree from Southeast University, China, in 2002. He is currently a chief professor and the dean of School of Information Engineering at Nanjing University of Finance and Economics. He is also a Ph.D. advisor of Nanjing University of Science and Technology. His main research interests include data mining, big data and e-commerce intelligence. He is the member of the ACM, CCF and IEEE Computer Society.
\end{IEEEbiography}\vspace{-1.3cm}

\begin{IEEEbiography}[{\includegraphics[width=1in,height=1.25in,clip,keepaspectratio]{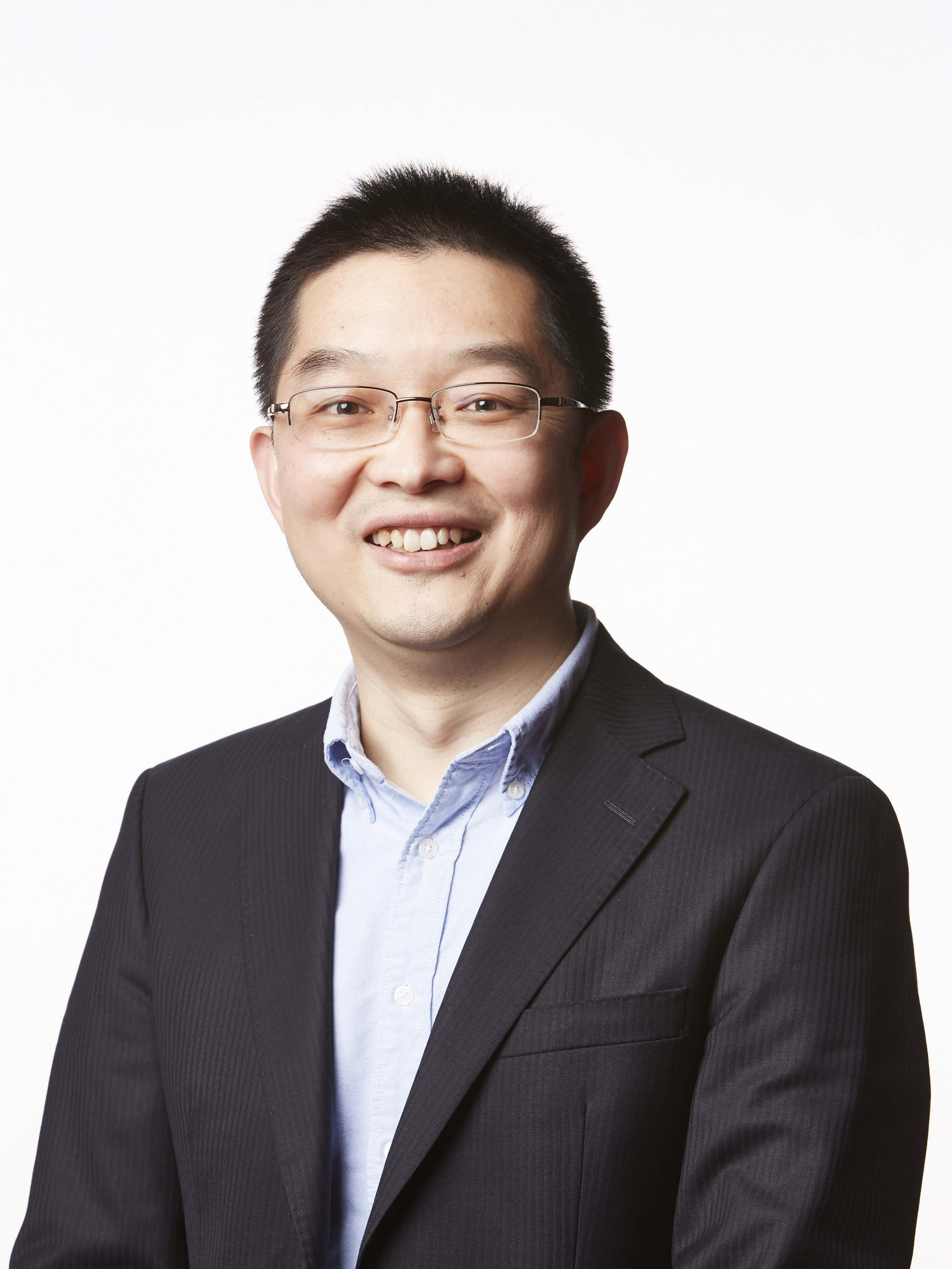}}]
{A. K. Qin} received the Ph.D. degree from Nanyang Technology University, Singapore, in 2007. Since 2013, he has been the Vice-Chancellor's Research Fellow, a Lecturer, and a Senior Lecturer with RMIT University, Melbourne, VIC, Australia. In 2017, he joined Swinburne University of Technology, Melbourne, VIC, Australia, as an Associate Professor. He is now leading Swinburne's Intelligent Data Analytics Lab as well as Machine Learning and Intelligent Optimization (MLIO) Research Group. His major research interests include evolutionary computation, machine learning, computer vision, GPU computing, and services computing.
\end{IEEEbiography}\vspace{-1.3cm}

\begin{IEEEbiography}[{\includegraphics[width=1in,height=1.25in,clip,keepaspectratio]{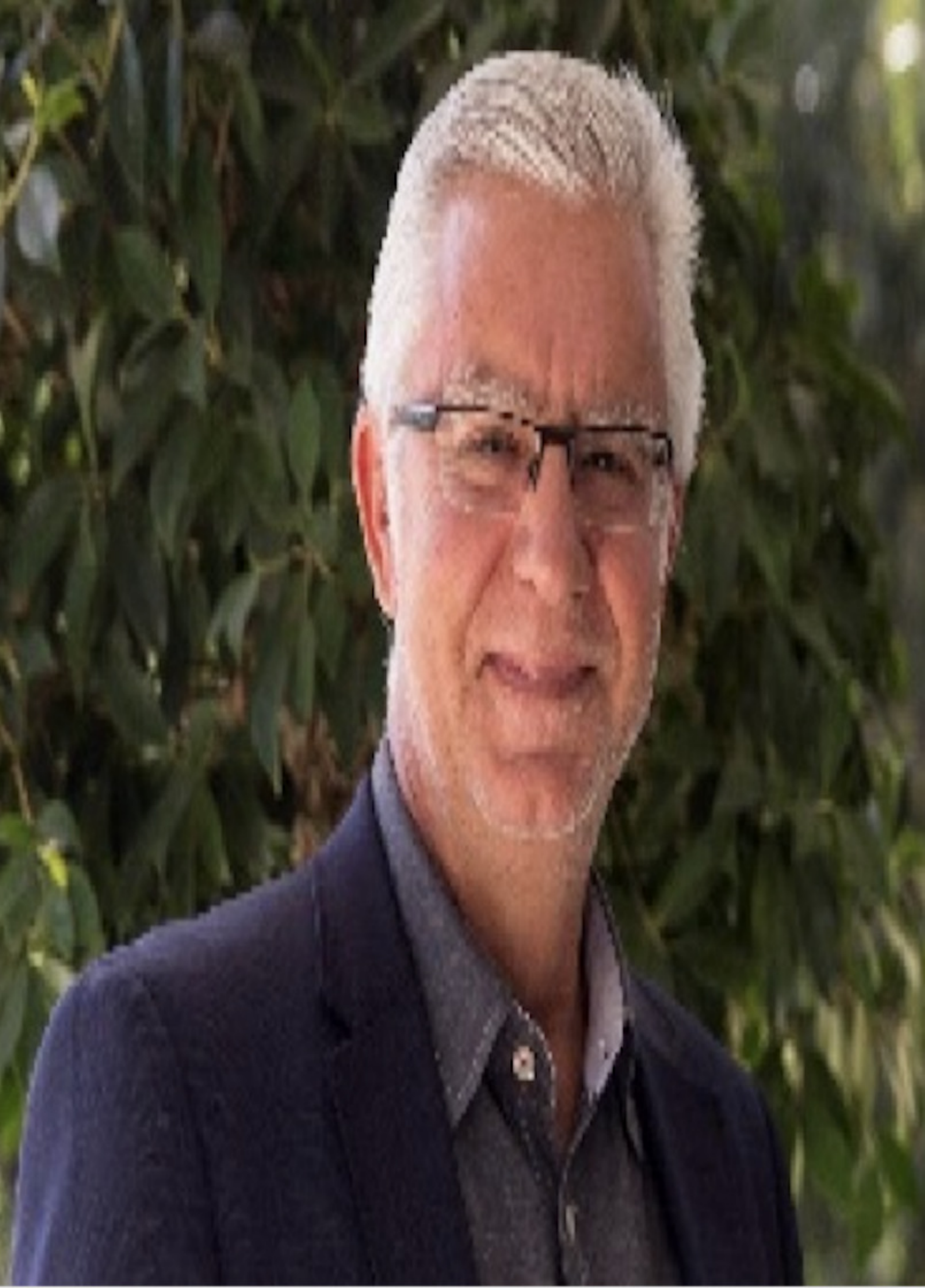}}]
{Timos Sellis} received the Ph.D. degree in computer science from the University of California, Berkeley, in 1986. He is a professor and director of the Data Science Research Institute at Swinburne University of Technology, Australia. Between 2013 and 2015, he was a professor at RMIT University, Australia, and before 2013 the director of the Institute for the Management of Information Systems (IMIS) and a professor at the National Technical University of Athens, Greece. His research interests include big data, data streams, personalization, data integration, and spatio-temporal database systems. He is a fellow of the IEEE and ACM. In 2018 he was awarded the IEEE TCDE Impact Award, in recognition of his impact in the field and for contributions to database systems research and broadening the reach of data engineering research.
\end{IEEEbiography}\vspace{-1.3cm}

\begin{IEEEbiography}[{\includegraphics[width=1in,height=1.25in,clip,keepaspectratio] {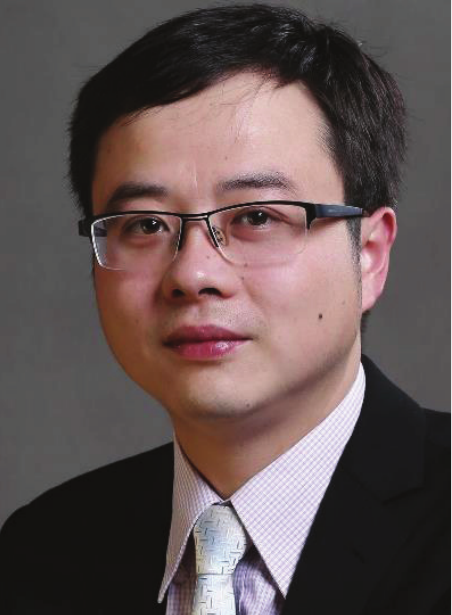}}]
{Zhiang Wu} (S'14-M'18) received his Ph.D. degree in Computer Science from Southeast University, China, in 2009. He is currently a full professor of School of Information Engineering at Nanjing University of Finance and Economics. His recent research focuses on distributed computing, data mining, e-commerce intelligence and social network analysis. He is the member of the ACM, IEEE and a senior member of CCF.
\end{IEEEbiography}

\end{document}